\pdfoutput=1
\documentclass[11pt]{article}

\usepackage{authblk}

\usepackage{emnlp2023}
\usepackage{times}
\usepackage{latexsym}

\usepackage[T1]{fontenc}
\usepackage[utf8]{inputenc}

\usepackage{microtype}

\usepackage{inconsolata}

\usepackage{lipsum}
\usepackage{yhmath}
\usepackage{amsfonts}\usepackage{hyperref}

\usepackage{amsmath}
\DeclareMathOperator*{\argmax}{arg\,max}
\DeclareMathOperator*{\softmax}{softmax}

\usepackage{graphicx}
\usepackage{calligra}
\usepackage{physics}
\usepackage{listings}
\usepackage{xcolor}
\usepackage{colortbl}
\usepackage{blindtext}
\usepackage{tikz}
\usepackage{caption}
\usepackage{subcaption} 
\usepackage{arydshln}
\usepackage{booktabs}
\definecolor{LightGray}{rgb}{0.93,0.93,0.93}
\definecolor{Fuchsia}{rgb}{0.6,0.0,0.5}
\definecolor{StrongGreen}{rgb}{0.0,0.7,0.1}
\definecolor{StrongRed}{rgb}{0.7,0.1,0.0}
\usepackage[colorinlistoftodos]{todonotes}
\usepackage{enumitem}
\usepackage{float}
\usepackage[normalem]{ulem}
\usepackage{calc}
\usepackage{multirow}

\newcommand\Ccancel{\bgroup\markoverwith
{\textcolor{red}{\rule[+0.5ex]{2pt}{0.8pt}}}\ULon}
\newcommand\Cu{\bgroup\markoverwith
{\textcolor{orange}{\rule[-0.5ex]{2pt}{0.8pt}}}\ULon}

\newcommand{\sm}[1]{\small#1\normalsize}
\newcommand{\tn}[1]{\scriptsize#1\normalsize}

\newcommand{\fyDone}[1]{\todo[color=cyan]{FY (done): #1}}
\renewcommand{\fyDone}[1]{}
\newcommand{\src}{\ensuremath{\mathbf{x}}}
\newcommand{\trg}{\ensuremath{\mathbf{y}}}
\newcommand{\simsrc}{\ensuremath{\mathbf{\tilde{x}}}} % source sentence
\newcommand{\simtrg}{\ensuremath{\mathbf{\tilde{y}}}} % target sentence
\newcommand{\mlevt}[1]{\texttt{TM$^{#1}$-LevT}}
\newcommand{\levt}[0]{\texttt{LevT}}

\newcommand{\tmlevt}[0]{\texttt{TM-LevT}}
\title{Towards Example-Based NMT with Multi-Levenshtein Transformers}

\newcommand{\systran}{\ensuremath{\clubsuit}} % \dag  \clubsuit
\newcommand{\isir}{\ensuremath{\heartsuit}} % \dag   \heartsuit

\author[$\isir$,$\systran$]{Maxime Bouthors}
\author[$\systran$]{Josep Crego}
\author[$\isir$]{Fran{\c c}ois Yvon}
\affil[$\isir$]{Sorbonne Université, CNRS, ISIR \\ F-75005 Paris, France}
\affil[$ $]{\texttt{firstname.lastname@isir.upmc.fr}}
\affil[$\systran$]{Systran \\ 5 rue Feydeau, F-75002 Paris, France}
\affil[$ $]{\texttt{firstname.lastname@systrangroup.fr}}

\begin{document}
\maketitle
\fyDone{Is Multi LevT appropriate as we do not have Levenshtein distances?}

\begin{abstract}
	\fyDone{Motivate / example-based / explication}
	Retrieval-Augmented Machine Translation (RAMT) is attracting growing attention. This is because RAMT not only improves translation metrics, but is also assumed to implement some form of domain adaptation. In this contribution, we study another salient trait of RAMT, its ability to make translation decisions more transparent by allowing users to go back to examples that contributed to these decisions.
	For this, we propose a novel architecture aiming to increase this transparency. This model adapts a retrieval-augmented version of the Levenshtein Transformer and makes it amenable to simultaneously edit multiple fuzzy matches found in memory. We discuss how to perform training and inference in this model, based on multi-way alignment algorithms and imitation learning. Our experiments show that editing several examples positively impacts translation scores, notably increasing the number of target spans that are copied from existing instances.
	% We present an extension of the edit-based Levenshtein Transformer with two main new assets. First, the ability to co-edit multiple sentences together. Second, we introduce a new paradigm of maximum-coverage optimality, replacing the edit-distance optimality one. This requires designing a new Dynamic Programming (DP) algorithm.
	% We also tackled the problem of misalignment occurring with multi-sentence editing.
	% We call this new model \textit{Multi-Levenshtein Transformer} and run experiments in a setting of
	% Example-Based Machine Translation (EBMT).
	% % Retrieval Augmented Neural Machine Translation (RANMT).
	% We showed that our model benefits from multiplicity, even when there is little diversity in the retrieved examples.
\end{abstract}

\fyDone{Plus l'accent sur la transparence}

\section{Introduction \label{sec:introduction}}

Neural Machine Translation (NMT) has become increasingly efficient and effective thanks to the development of ever larger encoder-decoder architectures relying on Transformer models \citep{vaswani-etal-2017-attention}. Furthermore, these architectures can readily integrate instances retrieved from a Translation Memory (TM) \citep{bulte-tezcan-2019-neural,xu-etal-2020-boosting,hoang-etal-2022-improving}, thereby improving the overall consistency of new translations compared to past ones. In such context, the autoregressive and generative nature of the decoder can make the process (a) computationally inefficient when the new translation has very close matches in the TM; (b) practically ineffective, as there is no guarantee that the output translation, regenerated from scratch, will resemble that of similar texts.

An alternative that is attracting growing attention is to rely on computational models tailored to edit existing examples and adapt them to new source sentences, such as the Levenshtein Transformer (\levt) model of \citet{gu-etal-2019-levenshtein}. This model can effectively handle \emph{fuzzy matches} retrieved from memories, performing minimal edits wherever necessary. As decoding in this model occurs is non-autoregressive, it is likely to be computationally more efficient. More important for this work, the reuse of large portions of existing translation examples is expected to yield translations that (a) are more correct; (b) can be transparently traced back to the original instance(s), enabling the user to inspect the edit operations that were performed. To evaluate claim (a) we translate our test data (details in Section~\ref{ssec:experimental-setting:data-and-metrics}) using a basic implementation of a retrieval-augmented \levt{} with TM (\tmlevt).
% without TM concatenation to the source \citep{xu-etal-2020-boosting}.
We separately compute the modified unigram and bigram precisions \citep{papineni-etal-2002-bleu} for tokens that are copied from the fuzzy match and tokens that are generated by the model.\footnote{Copies and generations are directly infered from the sequence of edit operations used to compute the output sentence.}
We observe that copies account for the largest share of output units and have better precision (see Table~\ref{table:mod-prec}).

\begin{table}[t]
	\centering
	\begin{tabular}{llrr}
		        &                    & precision & \% units \\ \hline
		unigram & \textit{copy}      & 87.5      & 64.9     \\
		        & \textit{gen}       & 52.6      & 35.1     \\ \hline
		bigram  & \textit{copy-copy} & 81.4      & 55.0     \\
		        & \textit{copy-gen}  & 40.1      & 8.9      \\
		        & \textit{gen-copy}  & 39.5      & 10.7     \\
		        & \textit{gen-gen}   & 34.2      & 25.4     \\
	\end{tabular}
	\caption{\label{table:mod-prec} Modified precision of copy vs. generated unigrams and bigrams for \tmlevt. For bigrams, we consider four cases: bigrams made of two copy tokens, two generated tokens, and one token of each type.}
\end{table}

Based on this observation, our primary goal is to optimize further the number of tokens copied from the TM. To do so, we propose simultaneously editing multiple fuzzy matches retrieved from memory, using a computational architecture -- Multi-LevT, or \mlevt{N} for short -- which extends \tmlevt{} to handle several initial translations. The benefit is twofold: (a) an increase in translation accuracy; (b) more transparency in the translation process.
Extending \tmlevt{} to \mlevt{N} however requires solving multiple algorithmic and computational challenges related to the need to compute Multiple String Alignments (MSAs) between the matches and the reference translation, which is a notoriously difficult problem; and designing appropriate training procedures for this alignment module.

Our main contributions are the following:
\begin{enumerate}
	\item a new variant of the \levt{} model that explicitly maximizes target coverage (\textsection{}\ref{ssec:optimal-alignment:solving});
	\item a new training regime to handle an extended set of editing operations (\textsection{}\ref{ssec:mlevt:roll-in-policy});
	\item two novel multiway alignment (\textsection{}\ref{ssec:optimal-alignment:solving}) and realignment (\textsection{}\ref{ssec:improving-mlevt}) algorithms;
	\item experiments in 11~domains where we observe an increase of BLEU scores, COMET scores, and the proportion of copied tokens (\textsection\ref{sec:experimental-results}).
\end{enumerate}

Our code and experimental configurations are available on github.\footnote{\url{https://github.com/Maxwell1447/fairseq/}}% will be released upon publication.

% ---
\section{Preliminaries / Background \label{sec:preliminary}}
\subsection{TM-based machine translation \label{ssec:preliminary:tmmt}}
Translation Memories, storing examples of past translations, is a primary component of professional Computer Assisted Translation (CAT) environments \citep{bowker-2002-computer}. Given a translation request for source sentence \src{}, TM-based translation is a two-step process: (a) retrieval of one or several instances $(\simsrc,\simtrg)$ whose source side resembles \src{}, (b) adaptation of retrieved example(s) to produce a translation. In this work, we mainly focus on step~(b), and assume that the retrieval part is based on a fixed similarity measure $\Delta$ between \src{} and stored examples. In our experiments, we use:
\begin{equation}
	\Delta(\src, \simsrc) = 1 - \frac{\operatorname{ED}(\src, \simsrc)}{\max(|\src|, |\simsrc|)},\label{eq:similarity}
\end{equation}
with $\operatorname{ED}(\src, \simsrc)$ the edit distance between \src{} and \simsrc{} and $|\src|$ the length of \src{}.
% First, sentences \simsrc{} with high similarity scores are retrieved in the hope that their target side will be close to the expected translation.
We only consider TM matches for which $\Delta$ exceeds a predefined threshold $\tau$
% \maxSmallTodo{$\alpha$ utilisé plus tard. Autre symbole ?}
and filter out the remaining ones. The next step, adaptation, is performed by humans with CAT tools. Here, we instead explore ways to perform this step automatically, as in Example-Based MT \citep{nagao-1984-framework,somers-1999-review,carl-etal-2004-recent}.
% \cite{Nagao,Sommers,Way}

\subsection{Adapting fuzzy matches with \levt \label{ssec:preliminary:adapting-fm}}
\begin{figure}[ht]
	\centering
	\includegraphics[scale=0.7]{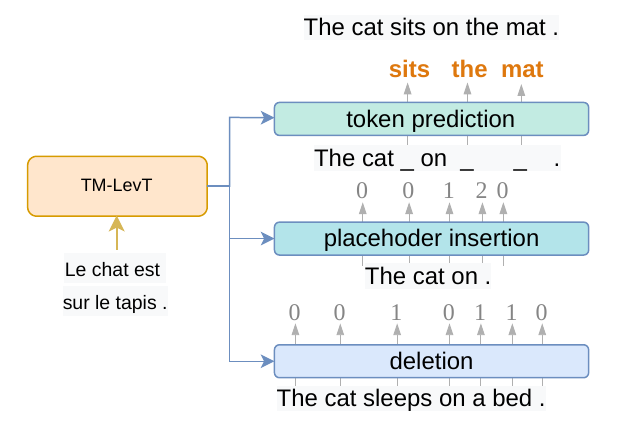}
	\caption{\label{figure:tm-levt} First decoding pass of \tmlevt, a variant of \levt{} augmented with Translation Memories.}
	\fyDone{the 20 is bizarre, check colors}
\end{figure}

The Levenshtein transformer of \citet{gu-etal-2019-levenshtein} is an encoder-decoder model which, given a source sentence, predicts edits that are applied to an initial translation in order to generate a revised output (Figure~\ref{figure:tm-levt}). The initial translation can either be empty or correspond to a match from a TM. Two editing operations -- insertion and deletion -- are considered. The former is composed of two steps: first, \emph{placeholder insertion}, which predicts the position and number of new tokens; second, the \emph{predictions of tokens} to fill these positions. Editing operations are applied iteratively in rounds of refinement steps until a final translation is obtained.

In \levt, these predictions rely on a joint encoding of the source and the current target and apply in parallel for all positions, which makes \levt{} a representative of non-autoregressive translation (NAT) models. As editing operations are not observed in the training data, \levt{} resorts to Imitation Learning, based on the generation of decoding configurations for which the optimal prediction is easy to compute.
Details are in \citep{gu-etal-2019-levenshtein}, see also \citep{xu-carpuat-2021-editor}, which extends it with a repositioning operation and uses it to decode with terminology constraints, as well as the studies of \citet{niwa-etal-2022-nearest} and \citet{xu-etal-2023-integrating} who also explore the use of \levt{} in conjunction with TMs.

\subsection{Processing multiple fuzzy matches \label{ssec:preliminary:processing-fm}}
One of the core differences between \mlevt{N} and \levt{} is its ability to handle multiple matches. This implies adapting the edit steps (in inference) and the roll-in policy (in imitation learning).

\paragraph{Inference in \mlevt{N} \label{paragraph:preliminary:processing-fm:sketch-infer}}
Decoding follows the same key ideas as for \levt{} (see Figure~\ref{figure:tm-levt}) but enables co-editing an arbitrary number $N$ of sentences. Our implementation (1) applies deletion, then placeholder insertion simultaneously on each retrieved example; (2) combines position-wise all examples into one single candidate sentence; (3) performs additional steps as in \levt: this includes first completing \textit{token prediction},  then performing \textit{Iterative Refinement} operations that edit the sentence to correct mistakes and improve it (\textsection\ref{ssec:mlevt:decoding}).
\fyDone{Would a graph help? Too late}

\paragraph{Training in \mlevt{N} \label{paragraph:preliminary:processing-fm:sketch-train}}
\mlevt{N} is trained with imitation learning and needs to learn the edit steps described above for both the first pass (1--2) and the iterative refinement steps (3). This means that we teach the model to perform the sequence of \textit{correct} edit operations needed to iteratively generate the reference output, based on the step-by-step reproduction of what an expert 'teacher' would do. For this, we need to compute the \textit{optimal operation} associated with each configuration (or state) (\textsection\ref{sec:optimal-alignment}). The \emph{roll-in} and \emph{roll-out policies} specify how the model is trained (\textsection\ref{ssec:mlevt:roll-in-policy}).

\section{Multi-Levenshtein Transformer \label{sec:mlevt}}

\subsection{Global architecture \label{ssec:mlevt:arch}}

\begin{figure}[t!]
	\centering
	\includegraphics[scale=0.85]{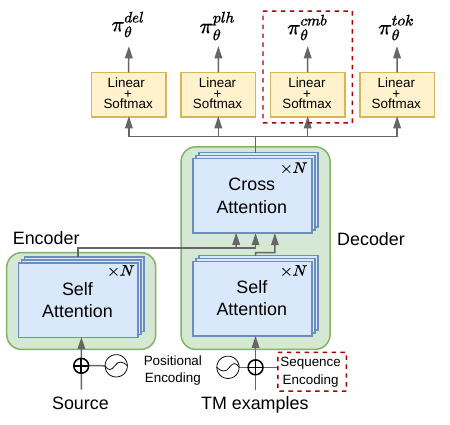}
	\caption{\label{figure:mlevt-architecture} A high-level overview of \mlevt{N}'s architecture. Additions w.r.t. \tmlevt{} are in a dashed box.}
\end{figure}

\mlevt{N} has two modes of operations: (a) the combination of multiple TM matches into one single sequence through alignment, (b) the iterative refinement of the resulting sequence. In step (a), we use the Transformer encoder-decoder architecture, extended with additional embedding and linear layers (see Figure~\ref{figure:mlevt-architecture}) to accommodate multiple matches. In each of the $N$ retrieved instances $\trg = (\trg_1 , \cdots , \trg_N )$, $\trg_{n,i}$ (the $i^{\text{th}}$ token in the $i^{\text{th}}$ instance) is encoded as
$E_{\trg_{n,i}} + P_i + S_n$, where $E \in \mathbb R^{| \mathcal V| \times d_{model}}$, $P \in \mathbb R^{L_{max} \times d_{model}}$ and $S \in \mathbb R^{(N + 1) \times d_{model}}$ are respectively the token, position and sequence embeddings. The sequence embedding identifies TM matches, and the positional encodings are reset for each $\trg_n$. The extra row in $S$ is used to identify the results of the combination and will yield a different representation for these single sequences.\footnote{
	$\trg^{op}$ denotes an intermediary sequence before applying edit operation $op$.
	$\trg^{op}_n \in \mathbb N^L$ is encoded with $S_n$; $\trg^{op}\in \mathbb N^L$ with $S_{N}$.}
% the entries when we have a single post-combination sequence, so the model considers this case differently (for $\trg^{tok}$ and refinement).
Once embedded, TM matches are concatenated and passed through multiple Transformer blocks,
% with cross-attention to the source $\src$,
until reaching the last layer, which outputs $(h_1, \cdots , h_{|\trg|})$ for a single input match or  $(h_{1,1},  \cdots, h_{1,|\trg_1|}, \cdots, h_{N,1}, \cdots , h_{N,|\trg_N |})$ in the case of multiple ones.
The \emph{learned policy} $\pi_\theta$ computes its decisions from these hidden states.
We use four classifiers, one for each sub-policy:

\begin{enumerate}
	\item \textit{deletion}: predicts \textit{keep} or \textit{delete} for each token $\trg_{n,i}^{del}$ with a projection matrix $A \in \mathbb R^{2 \times d_{model}}$:
	      \begin{align*}
		      \pi_{\theta}^{del}(d|n, i, & \trg_1^{del}, \cdots, \trg_N^{del}; \src) \\[-5pt]
		                                 & = \softmax \left( h_{n,i} A^T \right)
	      \end{align*}
	\item \textit{insertion}: predicts the number of placeholder insertions between $\trg_{n,i}^{plh}$ and $\trg_{n,{i+1}}^{plh}$ with a projection matrix $B \in \mathbb R^{(K_{max}+1) \times 2d_{model}}$:
	      \begin{align*}
		      \pi_\theta^{plh}(p|n, & i, \trg^{plh}_1, \cdots, \trg^{plh}_N; \src)                     \\[-5pt]
		                              & = \softmax \left( \left[ h_{n,i}, h_{n,i+1} \right] B^T \right),
	      \end{align*}
	      with $K_{max}$ the max number of insertions. %placeholders that can be added.
	\item \textit{combination}: predicts if token $\trg^{cmb}_{n,i}$ in sequence $n$ must be kept in the combination, with a projection matrix $C \in  \mathbb R^{2 \times d_{model}}$:
	      \begin{align*}
		      \pi_{\theta}^{cmb}(c|n, i, \trg^{cmb}_1, \cdots, \trg^{cmb}_N; \src) \\[-5pt]
		      = \softmax \left( h_{n,i} C^T \right).
	      \end{align*}
	\item \textit{prediction}: predicts a token in vocabulary $\mathcal V$ at each placeholder position, with a projection matrix $D \in \mathbb R^{|\mathcal V| \times d_{model}}$:
	      \[
		      \pi_\theta^{tok}(t|n, i, \trg^{tok}; \src) = \softmax \left( h_j D^T \right)
	      \]
\end{enumerate}
% \maxDone{Remplacé sigmoid par softmax, mais le retour à la ligne a tout cassé...}
Except for step~3, these classifiers are similar to those used in the original \levt.
%  note that in \mlevt{N} predictions are conditioned on all $N$ intermediary sequences (Figure~\ref{figure:mlevt-architecture}).\fyDone{why combination before pred ? why tok does not condition on all matches ?}

\subsection{Decoding \label{ssec:mlevt:decoding}}
% inference
% \input{decodinggraph}

\begin{figure}[!ht]
	\centering
	\includegraphics[scale=0.8]{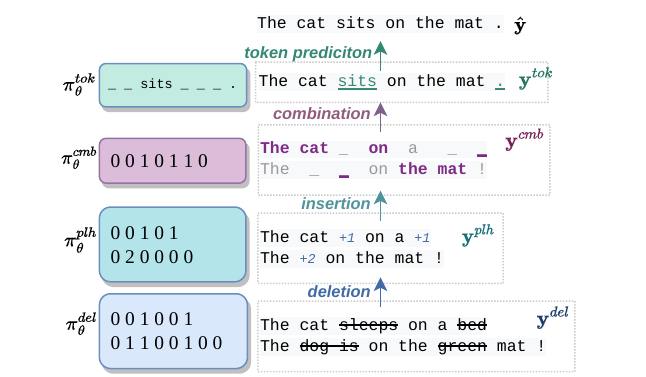}
	\caption{The first decoding pass in \mlevt{N}.}
	\label{fig:edition_steps_inference}
\end{figure}
\fyDone{colors to make the variants of y easier to read}

Decoding is an iterative process: in a first pass, the $N$ fuzzy matches are combined to compute a candidate translation; then, as in \levt, an additional series of \textit{iterative refinement} rounds \citep{gu-etal-2019-levenshtein} is applied until convergence or timeout. Figure~\ref{fig:edition_steps_inference} illustrates the first pass, where $N=2$ matches are first edited in parallel, then combined into one output.

To predict deletions (\textit{resp.} insertions and token predictions), we apply the \textit{argmax} operator to $\pi_\theta^{del}$ (\textit{resp.} $\pi_\theta^{plh}$, $\pi_\theta^{tok}$). For combinations, we need to aggregate separate decisions $\pi_\theta^{cmb}$ (one per token and match) into one sequence. For this, at each position, we pick the most likely token.
% at a given shared position.\fyTodo{Not clear}

% \fyTodo{A footnote}
During iterative refinement, we bias the model towards generating longer sentences since \levt{} outputs tend to be too short \citep{gu-etal-2019-levenshtein}. As in \levt, we add a penalty to the probability of inserting $0$ placeholder in $\pi_\theta^{plh}$
\citep{stern-etal-2019-insertion}. This only applies in the refinement steps to avoid creating more misalignments (see \textsection\ref{ssec:improving-mlevt}).

\subsection{Imitation learning \label{ssec:mlevt:roll-in-policy}}
\begin{figure*}[!ht]
	\centering
	\includegraphics[scale=0.9]{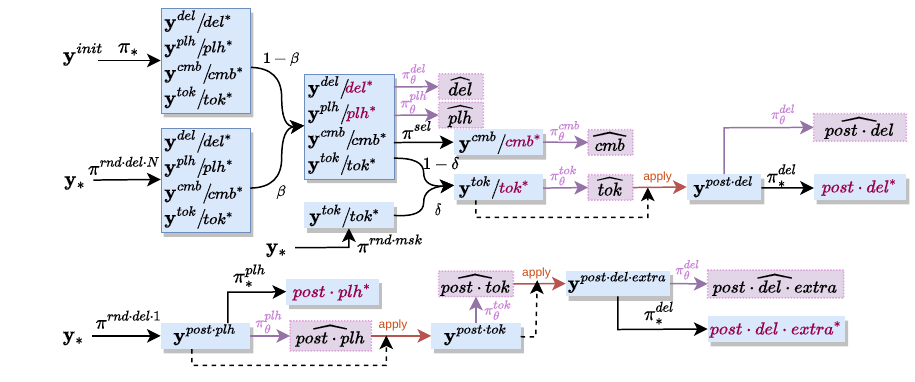}
	\caption{\label{figure:train-roll-in-graph} Roll-in policies used in training. Blue cells contain sets of target sentences (e.g.\ $\trg^{del}$), optionally associated with the expert prediction (e.g.\ $del^*$). Model's predictions are in Thistle and circumflexed (e.g.\ $\widehat{del}$). Pairs of model / expert predictions are summed in the loss: ($\widehat{del}$, $del^*$), ($\widehat{plh}$, $plh^*$), ($\widehat{cmb}$, $cmb^*$), ($\widehat{post \cdot del}$, $post \cdot del^*$), ($\widehat{post \cdot plh}$, $post \cdot plh^*$), ($\widehat{post \cdot del \cdot extra}$, $post \cdot plh \cdot extra^*$). "post" denotes policies applied in refinement steps.}

	% $\beta$ and $\delta$ are probabilities to use either of the sequences/targets (e.g. $y^{del}$/$del^*$). The final loss is the addition of cross entropy losses of output-target pairs ($\widehat{del}$, $del^*$), ($\widehat{plh}$, $plh^*$), ($\widehat{cmb}$, $cmb^*$), ($\widehat{post \cdot del}$, $post \cdot del^*$), ($\widehat{post \cdot plh}$, $post \cdot plh^*$), ($\widehat{post \cdot del \cdot extra}$, $post \cdot plh \cdot extra^*$). The targets used are in \textcolor{Fuchsia}{pink}. The label "post" denotes policies applied during the refinement steps.}
\end{figure*}

We train \mlevt{N} with Imitation Learning \cite{daume-etal-2009-search,ross-etal-2011-reduction},
teaching the system to perform the right edit operation for each decoding state. As these operations are unobserved in the training data, the standard approach is to simulate decoding states via a \emph{roll-in policy}; for each of these, the optimal decision is computed via an \emph{expert policy} $\pi_*$, composed of intermediate experts $\pi_*^{del}$, $\pi_*^{plh}$, $\pi_*^{cmb}$, $\pi_*^{tok}$.
% , which relies on a \emph{roll-out} policy to compute the reward of each possible choice.
The notion of optimality is discussed in \textsection{}\ref{sec:optimal-alignment}.
Samples of pairs (\emph{state}, \emph{decision}) are then used to train the system policy $\pi_\theta$.

First, from the initial set of sentences $\trg^{init}$, the unrolling of $\pi_*$ produces intermediate states $(\trg^{del}, del^*)$, $(\trg^{plh}, plh^*)$, $(\trg^{cmb}, cmb^*)$, $(\trg^{tok}, tok^*)$ (see top left in Figure~\ref{figure:train-roll-in-graph}).
Moreover, in this framework, it is critical to mitigate the \emph{exposure bias} and generate states that result from non-optimal past decisions \cite{zheng-etal-2023-towards}.
For each training sample $(\src, \trg_1, \cdots, \trg_N, \trg_*)$, we simulate multiple additional states as follows (see Figure~\ref{figure:train-roll-in-graph} for the full picture).
\begin{figure}[h]
	\centering
	\includegraphics[scale=0.84]{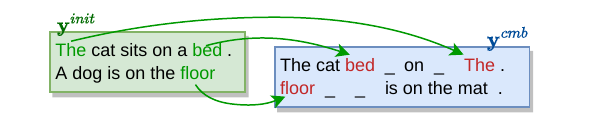}
	\caption{\label{figure:pi-sel} Noising $\trg^{cmb}$ with $\pi^{sel}$ using tokens from $\trg^{init} = (\trg_1, \cdots, \trg_N)$.}
\end{figure}
We begin with the operations involved in the first decoding pass:\footnote{Families of (\emph{state}, \emph{decision}) pairs that are novel with respect to \tmlevt{} are marked with $^\sharp$.}
\begin{enumerate}
	\item \textbf{Additional triplets}$^\sharp$: $\pi^{rnd \cdot del \cdot N}$ turns
	      % Instead of TM matches, we use $N$ random substrings of $\trg_*$ and teach the model to reinsert deleted tokens.
	      $\trg_*$ into $N$ random substrings, which simulates the edition of $N$ artificial examples.
	\item \textbf{Token selection}$^\sharp$ (uses $\pi^{sel}$): our expert policy never aligns two distinct tokens at a given position (\textsection\ref{ssec:optimal-alignment:deriving-edit-operations}). We simulate such cases that may occur at inference, as follows: with probability $\gamma$, each \verb|<PLH>| is replaced with a random token from fuzzy matches (Figure~\ref{figure:pi-sel}).
	      % This extra noise helps the model to only select relevant tokens. %  instead of just not placeholders.
\end{enumerate}
The expert always completes its translation in one decoding pass. Policies used in iterative refinement are thus trained with the following simulated states, based on roll-in and roll-out policies used in \levt{} and its variants
\citep{gu-etal-2019-levenshtein,xu-etal-2023-integrating,zheng-etal-2023-towards}:
\begin{enumerate}
	\setcounter{enumi}{2}
	\item \textbf{Add missing words} (uses $\pi^{rnd \cdot del \cdot 1}$):
	      % Start from a single subsequence $\trg_{post \cdot plh}$ of $\trg_*$ and teach to insert back the placeholders. 
	      with probability $\alpha$, $\trg^{post \cdot plh}$=$\trg_*$. With probability $1 - \alpha$, generate a subsequence $\trg^{post \cdot plh}$ with length sampled uniformly in $[0,\vert \trg_* \vert]$.
	      \label{item:pi-rnd-del-1}
	\item \textbf{Correct mistakes} (uses $\pi_\theta^{tok}$): using the output of token prediction $\trg^{post \cdot del}$, teach the model to erase the wrongly predicted tokens. \label{item:correct-mistakes}
	\item \textbf{Remove extra tokens}$^\sharp$ (uses $\pi_\theta^{ins}$, $\pi_\theta^{tok}$): insert placeholders in $\trg^{post \cdot tok}$ and predict tokens, yielding % from $\trg^{post \cdot plh}$,
	      $\trg^{post \cdot del \cdot extra}$, which trains the model to delete wrong tokens. These sequences differ from case~(\ref{item:correct-mistakes}) in the way \verb|<PLH>| are inserted.
	      \label{item:remove-wrong-extra}
	\item \textbf{Predict token}$^\sharp$ (uses $\pi^{rnd \cdot msk}$): each token in $\trg_*$ is replaced by \verb|<PLH>| with probability $\varepsilon$. As token prediction applies for both decoding steps, these states also improve the first pass.
\end{enumerate}

The expert decisions (e.g.\ inserting deleted tokens like in state~(\ref{item:pi-rnd-del-1}) ; or deleting wrongly predicted tokens in state~(\ref{item:correct-mistakes})) associated with most states are obvious, except for the initial state and state~(\ref{item:remove-wrong-extra}), which require an optimal alignment computation. % (\textsection\ref{sec:optimal-alignment}).

\section{Optimal Alignment \label{sec:optimal-alignment}}
Training the combination operation introduced above requires specifying the expert decision for each state.  While \levt{} derives its expert policy $\pi_*$ from the computation of \emph{edit distances}, we introduce another formulation based on the computation of \emph{maximal covers}.
For $N$=$1$, these formulations can be made equivalent\footnote{
	When the cost of \texttt{replace} is higher than \texttt{insertion} + \texttt{deletion}. This is the case in the original \levt{} code.
	% Only when \texttt{replace cost} $\geq$ \texttt{insertion + deletion costs}. This is the case in the orginal implementation of \levt.
} \citep{gusfield-dan-1997-algorithms}. %\done{rappeler $\pi_*$}
% \todo{Is is still relevant ? Important - may be a footnote would do?}
% \textit{roll-in} policy requires an \textit{optimal} decision in a given state. Whereas optimality is derived from \textit{edit distance} in LevT, we consider maximizing the target coverage.

\subsection{N-way alignments \label{ssec:optimal-alignment:formulation}}

\begin{figure*}
	\begin{subfigure}{0.49\textwidth}
		\centering
		\includegraphics[scale=0.94]{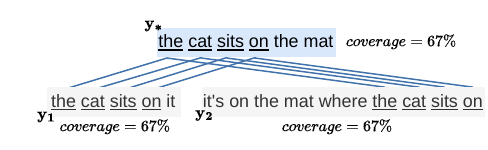}
		\caption{\label{figure:indep-align} Combination of the optimal 1-way alignments.}
	\end{subfigure}
	\begin{subfigure}{0.49\textwidth}
		\centering
		\includegraphics[scale=0.94]{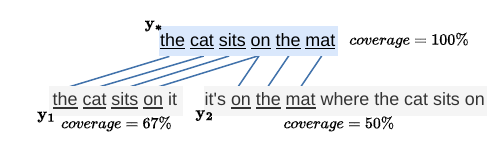}
		\caption{\label{figure:multi-align} Optimal 2-way alignment.}
	\end{subfigure}
	\caption{\label{figure:optimal-align} Illustration of the optimal N-way alignment which maximizes a global coverage criterion (\ref{figure:multi-align}), while independent alignments do not guarantee optimal usage of information present in TM examples (\ref{figure:indep-align}).}
\end{figure*}

We formulate the problem of optimal editing as an N-way alignment problem (see figure~\ref{figure:optimal-align}) that we define as follows. Given $N$ examples $(\trg_1, \cdots, \trg_N)$ and the target sentence $\trg_*$,
% , and retrieved conditionally to the source sentence $\src$,
a N-way alignment of $(\trg_1, \cdots, \trg_N)$ w.r.t. $\trg_*$ is represented as a bipartite graph $(V, V_*, E)$, where $V$ is further partitioned into $N$ mutually disjoint subsets $V_1 \dots V_N$. Vertices in each $V_n$ (resp.\ $V_*$) correspond to tokens in $\trg_n$ (resp.\ $\trg_*$). Edges $(n,i,j) \in E$ connect node $i$ in $V_n$ to node $j$ in $V_*$. An N-way alignment satisfies properties (i)-(ii):
\begin{enumerate}[label=(\roman*)]
	% \setlength\itemsep{-0.5em}
	% \item tokens in each $\trg_n$ can be linked to a token of $\trg_*$: $(n, i, j) \in \Gamma$ links $\trg_n^_{i}$ to $\trg_*^_{j}$
	\item Edges connect identical (matching) tokens: \newline $(n, i, j) \in E \Rightarrow \trg_{n,i} = \trg_{*,j} .$
	\item Edges that are incident to the same subset $V_n$ do not cross:\newline
	      $ (n, i, j), (n, i', j') \in E \Rightarrow (i'- i)(j' - j) > 0 .$
\end{enumerate}

% Such a graph indicates the reused tokens in the examples and determines what operations are needed to edit the examples into the target.
An optimal N-way alignment $E_*$ \textbf{maximizes the coverage} of tokens in $\trg_*$, then \textbf{the total number of edges}, where $\trg_{*,j}$ is covered if there exists at least one edge $(n, i, j) \in E$. Denoting $\mathbb E$ the set of alignments maximizing target coverage:
\begin{align*}
	 & \mathbb E =  \argmax_E \vert \{ \trg_{*,j}: \exists (n, i), (n, i, j) \in E \} \vert . \\
	 & E_* =        \argmax_{E \in \mathbb E} \vert E \vert .
\end{align*}

% Optimal alignments are closely related to edit distances. For $N$=$1$, maximizing the target coverage is equivalent to maximizing the common subsequence between $\trg_1$ and $\trg_*$, also equivalent to computing the basic edit distance\footnote{The basic edit distance only considers insertions and deletions, to which the Levenshtein distance adds substitutions.} \citep{gusfield-dan-1997-algorithms}.\fyTodo{Still relevant?}

\fyDone{This part comes later}
% Moreover, any edit solution can be represented as an alignment that unambiguously states the steps to edit the examples.

\subsection{Solving optimal alignment \label{ssec:optimal-alignment:solving}}

Computing the optimal N-way alignment is NP-hard (see Appendix~\ref{sec:appendix:NP-hard}).
% , requiring in a sense to test all the combinations of alignments of pairs $E_n \in \mathbb E_n$ (unless P=NP), where $\mathbb E_n$ is the set of maximal alignments in the pair $\trg_n$, $\trg_*$. Here the maximality has a specific meaning: an alignment $E_n$ is said to be maximal if, for any alignment $E_n'$, $E_n \subset E_n' \Rightarrow E_n = E_n'$. It implies that this is impossible to add a new edge in $E_n$.
This problem can be solved using Dynamic Programming (DP) techniques similar to Multiple Sequence Alignment (MSA) \citep{carrillo-etal-1988-multiple} with a complexity $O(N|\trg_*| \prod_n |\trg_n|)$. We instead implemented the following two-step heuristic approach:
% \fyTodo{Does it need revision ?}

\begin{enumerate}
	\item separately compute alignment graphs between each $\trg_n$ and $\trg_*$, then extract $k$-best 1-way alignments $\{E_{n, 1} \dots E_{n, k}\}$. This requires time $O(k|\trg_n||\trg_*|)$ using DP \citep{gusfield-dan-1997-algorithms};
	\item search for the optimal recombination of these graphs, selecting 1-way alignments $(E_{1,k_1} \dots E_{N,k_N})$ to form $E_* = \bigcup_n E_{n,k_n}$.
	      Assuming $N$ and $k$ are small, we perform an exhaustive search in $O(k^N)$.
\end{enumerate}

% Instead of designing a Dynamic Programming (DP) algorithm to find the exact solution, we decided to use a heuristic to decrease the complexity: we find the k-best candidates $E_n$ and find the combination $E_1, \cdots, E_N$ that maximizes the coverage. The k-best candidates correspond to a subset of $\mathbb E_n$ of size $k$, which contains the top-k coverages of $\trg_*$. These are obtained via a DP algorithm that finds the $k$-longest paths in a Directed Acyclic Graph (DAG). In the end, we run $N$ DP algorithm on each pair ($\trg_n$, $\trg_*$) instead of 1 global one.

\subsection{From alignments to edits \label{ssec:optimal-alignment:deriving-edit-operations}}

% From alignment $(V, V_*, E)$, we derive the optimal edits needed to compute $\trg_*$, and the associated intermediary sequences, applied in this order:
From an alignment $(V, V_*, E)$, we derive the optimal edits needed to compute $\trg_*$, and the associated intermediary sequences. Edges in $E$ indicate the tokens that are preserved throughout this process:
\begin{enumerate}
	\item \textit{deletion}: $\forall n, \forall i, \trg_{n,i}$ is kept only if $(n, i, j) \in E$ for some $j$; otherwise it is deleted. The resulting sequences are $\{\trg^{plh}_n\}_{n=1 \dots N}$.
	\item \textit{insertion}: Placeholders are inserted between successive tokens in all $\trg^{plh}_n$, resulting in the set $\{\trg^{cmb}_n\}_{n=1 \dots N}$, under the constraints that (a) all $\trg^{cmb}_n$ have the same length as $\trg_*$ and (b) non-placeholder tokens $\trg^{cmb}_{n,i}$ are equal to the reference token $\trg_{*,i}$.
	      % \maxSmallTodo{Pas vraiment... Soit on dit  $\trg^{cmb}_{n,i} = \trg^{cmb}_{m,i}$, soit que $\trg^{cmb}_{n,i}$ est relié à ${\trg_*}_i$}
	\item \textit{combination}: Sequences $\{\trg^{cmb}_n\}_{n=1 \dots N}$ are combined into $\trg^{tok}$ such that for each position $i$,
	      $\trg^{cmb}_{n,i} \neq \verb|<PLH>| \Rightarrow \trg^{tok}_{i} = \trg^{cmb}_{n,i}$. \\
	      If $\forall n, \trg^{cmb}_{n,i} = \verb|<PLH>|$,
	      then $\trg^{tok}_i = \verb|<PLH>|$.\fyDone{Improve pres}\fyDone{should be $\forall n$ ?}
	\item \textit{prediction}: The remaining \verb|<PLH>| symbols in $\trg^{tok}$ are replaced by the corresponding target token in $\trg_*$ at the same position.
\end{enumerate}

The \emph{expert policy} $\pi_*$ edits examples $\trg_1, \cdots, \trg_N$ into $\trg_*$ based on the optimal alignment $(V, V_*, E_*)$. It comprises $\pi_*^{del}$, $\pi_*^{plh}$, $\pi_*^{cmb}$, and $\pi_*^{tok}$, corresponding to the four steps listed above.

\section{Experiments}
\label{sec:experimental-setting}

\subsection{Data and metrics}
\label{ssec:experimental-setting:data-and-metrics}

We focus on translation from English to French and consider multiple domains.
This allows us to consider a wide range of scenarios, with a varying density of matching examples: our datasets include ECB, EMEA, Europarl, GNOME, JRC-Acquis, KDE4, PHP, Ubuntu, where high-quality matches are often available, but also News-Commentary, TED2013, and Wikipedia, where matches are more scarce (see Table~\ref{table:data-stats}, \textsection\ref{sec:appendix-data}).
\fyDone{On n'a pas mis les analyses ? Un tableau en annexe ?}

For each training sample $(\src, \trg)$, we retrieve up to 3 \emph{in-domain} matches. We filter matches $\src_n$ to keep only those with $\Delta(\src, \src_n)> 0.4$. We then manually split each of the 11 datasets into \emph{train}, \emph{valid}, \emph{test-0.4}, \emph{test-0.6}, where the \emph{valid} and \emph{test} sets contain 1,000 lines each.
% \textit{valid} also contains 1,000 pairs with no match found.
\textit{test-0.4} (resp.\ \textit{test-0.6}) contains samples whose best match is in the range $[0.4, 0.6[$ (resp.\ $[0.6, 1[$).
As these two test sets are only defined based on the \emph{best match score},  it may happen that some test instances will only retrieve 1 or 2 close matches (statistics are in Table~\ref{table:data-stats}).
% We found it relevant to separate the test set into two parts since lower \textit{fuzzy} scores imply that more examples could benefit, while high scores can already achieve already acceptable performances.

For the pre-training experiments (\textsection\ref{ssec:improving-mlevt}), we use a subsample of 2M random sentences from WMT'14. For all data, we use Moses tokenizer and 32k BPEs trained on WMT'14 with SentencePiece \citep{kudo-2018-subword}. We report BLEU scores \citep{papineni-etal-2002-bleu} and ChrF scores \citep{popovic-2015-chrf} as computed by \textit{SacreBLEU} \citep{post-2018-call}\fyDone{Use verb in sign.} and COMET scores \citep{rei-etal-2020-comet}. % in Appendix~\ref{sec:appendix-results}.

\subsection{Architecture and settings}
\label{ssec:experimental-setting:architecture-and-setting}

Our code\footnote{\url{https://github.com/Maxwell1447/fairseq}} extends Fairseq\footnote{\url{https://github.com/facebookresearch/fairseq}} implementation of \levt{} in many ways. It
% Our code is available at \url{
% 	???
% }.
% https://github.com/Maxwell1447/fairseq (insert for publication)
uses Transformer models \citep{vaswani-etal-2017-attention} (parameters in Appendix~\ref{sec:appendix-config}). Roll-in policy parameters (\textsection\ref{ssec:mlevt:roll-in-policy}) are empirically set as: $\alpha$=$0.3$, $\beta$=$0.2$, $\gamma$=$0.2$, $\delta$=$0.2$, $\varepsilon$=$0.4$. %without any exhaustive search.
% \todo{meta parameter search ?}
% \fyDone{Comparison of LevT and mLevT in terms of computation}
The AR baseline uses OpenNMT \cite{klein-etal-2017-opennmt} and uses the same data as \mlevt{} (Appendix~\ref{sec:appendix-config}).

\section{Results}
\label{sec:experimental-results}
\subsection{The benefits of multiple matches}
\label{ssec:experimental-results:benefit-multiple}
% court résultat sur 1-NN vs 2-NN

% ChrF (nrefs:1|case:mixed|eff:yes|nc:6|nw:0|space:no|version:2.1.0)

% | model            | 1    | 2    | 3    | all  |
% | ---------------- | ---- | ---- | ---- | ---- |
% | en-fr-1NN-normal | 63.6 | 65.0 | 68.4 | 66.8 |
% | en-fr-3NN-normal | 64.1 | 65.8 | 69.3 | 67.5 |

\begin{table}[h]
	\centering
	\scalebox{1}{
		% \begin{tabular}{lrrrr}
		\begin{tabular}{p{0.09\textwidth}>{\centering}p{0.06\textwidth}>{\centering}p{0.06\textwidth}>{\centering\arraybackslash}p{0.06\textwidth}>{\centering\arraybackslash}p{0.06\textwidth}}
			Model \textbackslash N      & 1                   & 2                   & 3                   & all                 \\ \hline
			% \rowcolor{LightGray}
			size                        & \sm{4,719}          & \sm{2,369}          & \sm{14,912}         & \sm{22,000}         \\
			\hdashline
			\multirow{ 2}{*}{\mlevt{1}} & \sm{45.8}/\tn{63.6} & \sm{48.7}/\tn{65.0} & \sm{55.0}/\tn{68.4} & \sm{52.0}/\tn{66.8} \\
			                            & \sm{19.6}           & \sm{26.2}           & \sm{41.5}           & \sm{35.0}           \\
			\multirow{ 2}{*}{\mlevt{3}} & \sm{46.6}/\tn{64.1} & \sm{50.0}/\tn{65.8} & \sm{56.0}/\tn{69.3} & \sm{53.0}/\tn{67.5} \\
			                            & \sm{14.0}           & \sm{16.0}           & \sm{38.2}           & \sm{30.8}           \\
		\end{tabular}
	}
	\caption{\label{table:benefit-N} BLEU/\sm{ChrF} and COMET scores on the full test set. All BLEU/\sm{ChrF} differences are significant ($p=0.05$).}
\end{table}
% \maxSmallTodo{\textbf{all} is BLEU on concatenation, whereas in the two big tables, it is the mean of BLEU in domains... Où préciser ça??}
\fyDone{Add COMET scores in this one}\fyDone{Add significance -> all significant...}
\fyDone{Comment COMET score vs BLEU (BP?)}

We compare two models in Table~\ref{table:benefit-N}: one trained with one TM match, the other with three. Each model is evaluated with, at most, the same number of matches seen in training. This means that \mlevt{1} only uses the 1-best match, even when more examples are found. In this table, test sets \textit{test-0.4} and \textit{test-0.6} are concatenated, then partitioned between samples for which exactly 1, 2, and 3 matches are retrieved.
We observe that \mlevt{3}, trained with 3 examples, consistently achieves better BLEU and ChrF scores than \mlevt{1}, even in the case $N$=1, where we only edit the closest match.\footnote{This is because the former model has been fed with more examples during training, which may help regularization.} These better BLEU scores are associated with a larger number of copies from the retrieved instances, which was our main goal (Table~\ref{table:origin-proportion}). Similar results for the other direction are reported in the appendix \textsection~\ref{sec:reverse-dir}  (Table~\ref{tab:benefit-N-reverse-dir}).

\begin{table}[htbp]
	\centering
	\scalebox{0.8}{
		\begin{tabular}{llrrr}
			        &                    & {\tmlevt} & {\mlevt{1}} & {\mlevt{3}} \\ \hline
			unigram & \textit{copy     } & 64.9      & 64.5        & 68.8        \\
			        & \textit{gen      } & 35.1      & 35.5        & 31.2        \\ \hline
			bigram  & \textit{copy-copy} & 55.0      & 54.5        & 58.0        \\
			        & \textit{copy-gen } & 8.9       & 9.0         & 10.1        \\
			        & \textit{gen-copy } & 10.7      & 10.8        & 11.0        \\
			        & \textit{gen-gen  } & 25.4      & 25.7        & 20.9        \\
		\end{tabular}
	}
	\caption{\label{table:origin-proportion} Proportion of unigrams and bigram from a given origin (copy vs. generation) for various models.}
\end{table}

We report the performance of systems trained using $N$=$1,2,3$ for each domain and test set in Table~\ref{table:BLEU-num-example} (BLEU) and \ref{table:comet-num-example} (COMET). We see comparable average BLEU scores for $N$=1 and $N$=3, with large variations across domains, from which we conclude that: (a) using 3 examples has a smaller return when the best match is poor, meaning that bad matches are less likely to help (\textit{test-0.4} vs. \textit{test-0.6}); (b) using 3 examples seems advantageous for narrow domains, where training actually exploits several close matches (see also Appendix~\ref{sec:appendix-results}). We finally note that COMET scores\footnote{Those numbers are harder to interpret, given the wide range of COMET scores across domains (from $\approx$ -40 to +86).} for \mlevt{3} are always slightly lower than for \mlevt{1}, which prompted us to develop several extensions.

% There is no concatenation of the examples to the source, which makes the results hardly comparable to other works using a similar setup \citep{bulte-tezcan-2019-neural,xu_boosting_2020}. --> move to limitations.

\begin{table*}[ht]
	\centering
	\scalebox{0.79}{
		\begin{tabular}{llrrrrrrrrrrr|r}
			\hline
			\rowcolor{LightGray}
			                  &                                   & \makebox[\widthof{~ECB~}]{ECB} & \makebox[\widthof{~ECB~}]{EME} & \makebox[\widthof{~ECB~}]{Epp} & \makebox[\widthof{~ECB~}]{GNO} & \makebox[\widthof{~ECB~}]{JRC} & \makebox[\widthof{~ECB~}]{KDE} & \makebox[\widthof{~ECB~}]{News} & \makebox[\widthof{~ECB~}]{PHP} & \makebox[\widthof{~ECB~}]{TED} & \makebox[\widthof{~ECB~}]{Ubu} & \makebox[\widthof{~ECB~}]{Wiki} & \textbf{all}  \\ \hline
			\textit{test-0.4} & AR                                & \textbf{63.0}                  & \textbf{63.6}                  & \textbf{43.6}                  & \textbf{69.3}                  & \textbf{75.1}                  & \textbf{62.8}                  & \textbf{28.8}                   & \textbf{41.2}                  & \textbf{42.2}                  & \textbf{59.1}                  & \textbf{42.2}                   & \textbf{55.8} \\ \hdashline
			                  & \tmlevt                           & \textit{50.5}                  & \textit{50.7}                  & \textit{31.3}                  & \textit{54.3}                  & \textit{62.4}                  & \textit{47.9}                  & \textit{18.0}                   & \textit{30.1}                  & \textit{24.2}                  & \textit{43.3}                  & \textit{29.8}                   & \textit{42.8} \\[-0.5pt] \hdashline

			                  & \mlevt{1}                         & 53.1                           & 53.7                           & 35.5                           & 60.3                           & 65.6                           & 51.8                           & 22.2                            & 31.7                           & 30.2                           & 48.8                           & 32.0                            & 46.2          \\[-0.5pt] \hdashline
			                  & \mlevt{2}                         & \textbf{54.0}                  & 54.3                           & \textit{34.0}                  & 60.5                           & 66.0                           & 53.2                           & \textit{20.7}                   & \textbf{33.7}                  & \textit{28.9}                  & 48.0                           & 32.6                            & 46.5          \\[-0.5pt] \hdashline
			                  & \mlevt{3}                         & \textbf{53.9}                  & \textbf{55.6}                  & \textit{34.2}                  & 60.8                           & 66.0                           & \textbf{53.5}                  & \textit{20.4}                   & \textbf{33.1}                  & \textit{28.6}                  & \textit{47.5}                  & 32.9                            & \textbf{46.5} \\[-0.5pt] \hdashline
			                  & \multicolumn{1}{l}{~~~+pre-train} & \textbf{54.9}                  & \textbf{55.9}                  & \textit{34.4}                  & \textbf{62.7}                  & \textbf{67.4}                  & \textbf{54.1}                  & \textit{21.1}                   & \textbf{34.7}                  & 30.1                           & 49.3                           & \textbf{33.5}                   & \textbf{47.5} \\[-0.5pt] \hdashline
			                  & \multicolumn{1}{l}{~~~+realign}   & \textbf{54.4}                  & \textbf{55.9}                  & \textit{34.4}                  & \textbf{61.2}                  & 66.2                           & 53.2                           & \textit{20.4}                   & \textbf{33.3}                  & \textit{28.4}                  & 47.9                           & \textbf{33.1}                   & \textbf{46.7} \\[-0.5pt] \hdashline
			                  & \multicolumn{1}{l}{~~~+both}      & \textbf{55.0}                  & \textbf{56.0}                  & 34.9                           & \textbf{62.8}                  & \textbf{67.5}                  & \textbf{54.0}                  & \textit{21.4}                   & \textbf{34.8}                  & 30.8                           & 49.6                           & \textbf{33.9}                   & \textbf{47.8} \\[-0.5pt]

			\hline\hline
			\textit{test-0.6} & AR                                & \textbf{69.7}                  & \textbf{70.4}                  & \textbf{57.4}                  & \textbf{80.6}                  & \textbf{82.4}                  & \textbf{68.2}                  & \textbf{26.1}                   & \textbf{46.4}                  & \textbf{62.5}                  & \textbf{68.5}                  & \textbf{68.7}                   & \textbf{66.6} \\ \hdashline
			                  & \tmlevt                           & \textit{59.0}                  & 64.0                           & \textit{45.8}                  & \textit{66.9}                  & \textit{73.5}                  & \textit{53.4}                  & \textit{18.8}                   & \textit{34.7}                  & \textit{49.1}                  & \textit{53.2}                  & \textit{58.9}                   & \textit{55.8} \\[-0.5pt] \hdashline

			                  & \mlevt{1}                         & 60.5                           & 64.6                           & 48.9                           & 69.7                           & 75.7                           & 57.2                           & 21.0                            & 36.2                           & 55.0                           & 58.3                           & 62.2                            & 58.2          \\[-0.5pt] \hdashline
			                  & \mlevt{2}                         & \textbf{62.7}                  & \textbf{67.0}                  & \textbf{50.0}                  & \textbf{71.7}                  & 76.2                           & \textbf{60.2}                  & 21.7                            & \textbf{38.6}                  & 54.2                           & \textbf{59.8}                  & 62.8                            & \textbf{59.7} \\[-0.5pt] \hdashline
			                  & \mlevt{3}                         & \textbf{63.8}                  & \textbf{67.4}                  & \textbf{50.0}                  & \textbf{71.1}                  & \textbf{76.4}                  & \textbf{60.0}                  & 21.5                            & \textbf{39.2}                  & 54.3                           & \textbf{59.6}                  & 62.3                            & \textbf{60.0} \\[-0.5pt] \hdashline
			                  & \multicolumn{1}{l}{~~~+pre-train} & \textbf{64.9}                  & \textbf{68.3}                  & \textbf{50.3}                  & \textbf{72.7}                  & \textbf{77.3}                  & \textbf{62.3}                  & \textbf{21.8}                   & \textbf{40.7}                  & 54.6                           & \textbf{61.3}                  & \textbf{65.0}                   & \textbf{61.1} \\[-0.5pt] \hdashline
			                  & \multicolumn{1}{l}{~~~+realign}   & \textbf{64.0}                  & \textbf{68.0}                  & \textbf{50.2}                  & \textbf{71.5}                  & \textbf{76.5}                  & \textbf{59.9}                  & 21.6                            & \textbf{39.0}                  & 54.7                           & \textbf{60.0}                  & 63.1                            & \textbf{60.2} \\[-0.5pt] \hdashline
			                  & \multicolumn{1}{l}{~~~+both}      & \textbf{65.0}                  & \textbf{68.3}                  & \textbf{50.8}                  & \textbf{73.7}                  & \textbf{77.4}                  & \textbf{62.3}                  & \textbf{22.0}                   & \textbf{40.6}                  & 54.7                           & \textbf{61.4}                  & \textbf{65.3}                   & \textbf{61.3} \\[-0.5pt]
		\end{tabular}
	}
	\caption{\label{table:BLEU-num-example} Per domain BLEU scores for \tmlevt, \mlevt{N} and variants. Bold (resp.\ italic) for scores significantly higher (resp. lower) than \mlevt{1} ($p=0.05$). $p$-values from SacreBLEU paired bootstrap resampling $(n=1000)$. The Autoregressive (AR) system is our implementation of \citep{bulte-tezcan-2019-neural}.}
	% Italic results are less than 0.5 BLEU under \mlevt{1}.}
\end{table*}

\subsection{Improving \mlevt{N}}
\label{ssec:improving-mlevt}

\paragraph{Realignment}
\label{paragraph:realignment}
\begin{figure}[ht]
	\centering
	\includegraphics[scale=0.54]{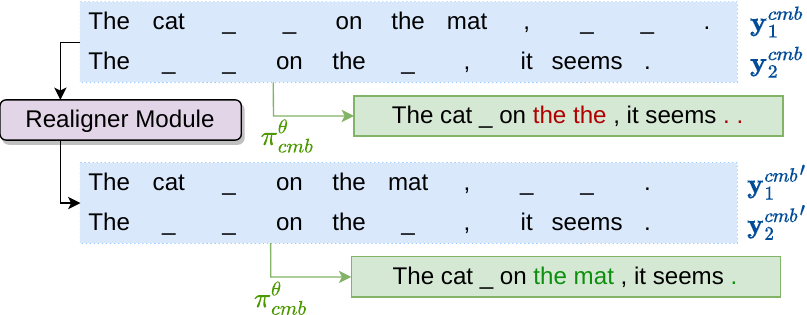}
	% \includesvg[scale=0.35]{realigner-example.drawio.svg}
	\caption{\label{figure:realigner-illustration} Fixing misalignments with realignment}
\end{figure}
In preliminary experiments, we observed that small placeholder prediction errors in the first decoding pass could turn into catastrophic misalignments
% , resulting in strange combinations of sentences in $\trg^{tok}$
(Figure~\ref{figure:realigner-illustration}). To mitigate such cases, we introduce an additional realignment step during inference, where some predicted placeholders are added/removed if this improves the global alignment. Realignment is formulated as an optimization problem aimed to perform a tradeoff between the score $- \log \pi_\theta^{plh}$ of placeholder insertion and an alignment cost (see Appendix~\ref{sec:appendix-realignment}).
% Since our DP implementation turned out to be either extremely memory-expensive or time expensive, we alternatively designed a gradient descent algorithm where the numbers of placeholders we can insert are turned into continuous values we optimize. See for more details.

We assess realignment for $N$=$3$ (Tables~\ref{table:BLEU-num-example}  and \ref{table:comet-num-example}) and observe small, yet consistent average gains (+0.2 BLEU, +1.5 COMET) for both test sets. % , with variations across domains.
% it improves \texttt{mTlev-3} by +0.5 BLEU on \textit{test-0.4} and +1.0 BLEU on \textit{test-0.6}, also outperforming the average score of \mlevt{1} overall (+0.6 on average on both test sets). %.

% (in \textit{test-0.4}: JRC-Acquis, NewsCommentary, TED2013 and Ubuntu; in \textit{test-0.6}: GNOME, JRC-Acquis, KDE4 and Ubuntu). We assume these datasets are prone to redundancy in the retrieved examples, thus making multiple matches often irrelevant.
% \maxSmallTodo{Pareil (type de données) !}

\paragraph{Pre-training}
\label{paragraph:pre-trained}
Another improvement uses pre-training with synthetic data. For each source/target pair $(\src, \trg)$ in the pre-training corpus, we simulate $N$ fuzzy matches by extracting from \trg{}
% evenly distributed
$N$ substrings $\trg_n$ of length $\approx \vert \trg \vert \cdot r$, with $r \in [0,1]$. Each $\trg_n$ is then augmented as follows:

\begin{enumerate}
	% \setlength\itemsep{0em}
	% \item We set $\trg_n'$ to be the chunk subsequence $\trg^{[
	% 				      \lfloor (1-r) \vert \trg \vert \frac{1-n}{1-N}\rfloor :
	% 				      \lfloor \vert \trg \vert \left(r + (1-r)\frac{1-n}{1-N} \right)\rfloor
	% 			      ]}$
	%       whose size is approximatively $\vert \trg \vert \cdot r$.
	%       \maxSmallTodo{a-t-on besoin de cette formule dégoûtante ?}
	\item We randomly insert placeholders to increase the length by a random factor between $1$ and $1+f$, $f=0.5$ in our experiments.
	      %	\item We insert mask tokens so that the new sequence length is uniformly distributed between $\vert {\trg_n} \vert$ and $(1 + f) * \vert {\trg_n} \vert$, with $f>0$ a size factor ($0.5$ in our experiments).
	\item We use the CamemBERT language model \citep{martin-etal-2020-camembert} to fill the masked tokens. % non-autoregressively
	      %	\item We finally shuffle the resulting sequences.
\end{enumerate}
These artificial instances
% consisting of $(N+2)$-uplets $(\src, \trg_1, \cdots, \trg_N, \trg)$
simulate diverse fuzzy matches and are used to pre-train a model, using the same architecture and setup as in \textsection\ref{ssec:experimental-setting:architecture-and-setting}.
% This is condition \texttt{+pre-train} in Table~\ref{table:BLEU-num-example}.
Pre-training yields markedly higher scores than the baseline (+1.3 BLEU, +6.4 COMET for \textit{test-0.4} and +0.9 BLEU, +4.6 COMET for \textit{test-0.6}). Training curves also suggest that pre-trained models are faster to converge. Combining with realignment yields additional gains for \mlevt{3}, which \emph{outperforms} \mlevt{1} \emph{in all domains and both metrics}.
%\fyDone{Do we have +realign in inference for this one ?}

\paragraph{Knowledge distillation}
\label{paragraph:kd}
\textit{Knowledge Distillation} (KD) \citep{kim-rush-2016-sequence} is used to mitigate the effect of multimodality of NAT models \citep{zhou-2021-understanding} and to ease the learning process. We trained a \mlevt{N} model with distilled samples $(\src, \tilde\trg_1, \cdots, \tilde\trg_N, \tilde\trg)$, where automatic translations $\tilde\trg_i$ and $\tilde\trg$ are derived from their respective source $\src_i$ and $\src$ with an auto-regressive teacher trained with a concatenation of all the training data.
% computed from $( \src_1, \cdots, \src_N, \src)$. $(\src_1, \cdots, \src_N)$ are the source-side fuzzy matches.

We observe that KD is beneficial (+0.3 BLEU) for low-scoring matches (\textit{test-0.4}) but hurts performance (-1.7 BLEU) for the better ones in \textit{test-0.6}. This may be because the teacher model, with a BLEU score of $56.7$ on the \textit{test-0.6}, fails to provide the excellent starting translations the model can access when using non-distilled data.\fyDone{Shorten, cite v ?}
% It is most likely because the model learned to be less conservative than it should be for high-scoring matches.

% \begin{table}[!ht]
% 	\centering
% 	\begin{tabular}{llrr}
% 		model     &              & \textit{test-0.4} & \textit{test-0.6} \\ \hline
% 		\mlevt{3} &              & 46.5              & \textbf{60.0}     \\[-1pt]
% 		          & \textit{+KD} & \textbf{46.8}     & 58.3              \\
% 	\end{tabular}
% 	\caption{\label{table:kd-short} BLEU scores for the distilled model.}
% \end{table}
% \vspace{-\baselineskip}

\subsection{Ablation study}
\label{ssec:experimental-results:analysis}

% \subsubsection{Analysis}
% \paragraph{Ablation study}
We evaluate the impact of the various elements in the mixture roll-in policy via an ablation study (Table~\ref{table:ablation-study}). Except for $\pi^{sel}$, every new element in the roll-in policy increases performance. As for $\pi^{sel}$, our system seems to be slightly better with than without. An explanation is that, in case of misalignment, the model is biased towards selecting the first, most similar example sentence.\fyDone{Do not we shuffle ?} As an ablation, instead of aligning by globally maximizing coverage (\textsection~\ref{ssec:optimal-alignment:solving}), we also compute alignments that maximize coverage independently as in figure~\ref{figure:indep-align}.\fyDone{Is it necessary?}

\begin{table}[h]
	\centering
	\scalebox{0.8}{
		\begin{tabular}{llrr}
			          &                       & \textit{~~test-0.4} & \textit{~~test-0.6} \\ \hline
			\mlevt{3} &                       & \textbf{46.5}       & \textbf{60.1}       \\[-1pt]
			          & \texttt{-sel    }     & 46.2                & 60.0                \\[-1pt]
			          & \texttt{-delx   }     & 44.8                & 58.6                \\[-1pt]
			          & \texttt{-rnd-del }    & 38.6                & 51.9                \\[-1pt]
			          & \texttt{-rnd-mask}    & 46.0                & 59.0                \\[-1pt]
			          & \texttt{-dum-plh}     & 41.0                & 50.9                \\[-1pt]
			          & \texttt{-indep-align} & 42.6                & 56.4                \\[-1pt]
		\end{tabular}
	}
	\caption{\label{table:ablation-study-short} Ablation study. We build models with variable roll-in policies: \texttt{-sel}: no random selection noise ($\gamma$=0); \texttt{-delx}: no extra deletion;\texttt{-rd-del}: no random deletion ($\beta$=0); \texttt{-mask}: no random mask ($\delta$=0); \texttt{-dum-plh}: never start with $\trg_{post \cdot del}$=$\trg_*$ ($\alpha$=0); \texttt{-indep-align}: alignments are independent. Full results in Appendix~\ref{sec:appendix-results}.}
\end{table}

% \paragraph{Comparing alignments} Our experiments include two models with $N$=$1$, the only difference being the way alignments underlying the expert policy are computed: either minimizing the edit-distance (\levt) or maximizing the coverage (\mlevt{1}).
% We also compared to a model using the original roll-in policy of Levenshtein Transformer, with the only difference being an initial deletion on the examples (\textit{LevT-1 original}).
% Indeed, the original code available is for generation only.
% Referring again to Table~\ref{table:BLEU-num-example}, we note that our new alignment paradigm clearly outperforms the baseline.
% \fyDone{What the point of next text (commented)?}
% , even though the type of retrieval (here \textit{greedy fuzzy match}) does not favor diversity and contains a lot of redundancy.

A complete run of \mlevt{N} is in Appendix~\ref{sec:appendix-results}. % (Table~\ref{table:illustration-php-04-571}).

\section{Related Work \label{sec:related}}

As for other Machine Learning applications, such as text generation \citep{guu-etal-2018-generating}, efforts to integrate a retrieval component in neural-based MT have intensified in recent years. One motivation is to increase the transparency of ML models by providing users with tangible traces of their internal computations in the form of retrieved examples \citep{rudin-cynthia-2019-stop}. For MT, this is achieved by integrating fuzzy matches retrieved from memory as an additional conditioning context. This can be performed simply by concatenating the retrieved target instance to the source text \cite{bulte-tezcan-2019-neural}, an approach that straightforwardly accommodates several TM matches \cite{xu-etal-2020-boosting}, or the simultaneous exploitation of their source and target sides \cite{pham-etal-2020-priming}. More complex schemes to combine retrieved examples with the source sentence are in \citep{gu-etal-2018-search,xia-etal-2019-graph,he-etal-2021-fast}. % \citet{He21fast}, which encodes the TM match using the decoder embedding matrix and performs a cross-attention between the decoder input and the encoded TM match
% Fast and accurate neural machine translation with translation memory
% \maxSmallTodo{he-etal-2020-fast ou He21fast introuvable ????}
The recent work of \citet{cheng-etal-2022-neural} handles multiple complementary TM examples retrieved in a \emph{contrastive manner} that aims to enhance source coverage.
% In their architecture, similar examples are used in an additional "TM attention" layer that is located on top of the self- and cross-attention layers; their representations are computed through a dedicated hierarchical encoder.
\citet{cai-etal-2021-neural} also handle multiple matches and introduce two novelties: (a) retrieval is performed in the target language and (b) similarity scores are trainable, which allows to evaluate retrieved instances based on their usefulness in translation. Most of these attempts rely on an auto-regressive (AR) decoder, meaning that the impact of TM match(es) on the final output is only indirect.

The use of TM memory match with a NAT decoder is studied in \citep{niwa-etal-2022-nearest,xu-etal-2023-integrating,zheng-etal-2023-towards}, which adapt \levt{} for this specific setting, using one single retrieved instance to initialize the edit-based decoder. Other evolutions of \levt, notably in the context of constraint decoding, are in \citep{susanto-etal-2020-lexically,xu-carpuat-2021-editor}, while a more general account of NAT systems is in \citep{xiao-etal-2022-survey}.

\citet{zhang-etal-2018-guiding} explore a different set of techniques to improve translation using retrieved segments instead of full sentences. Extending KNN-based language models \citep{he-etal-2021-efficient} to the conditional case, \citet{khandelwal_nearest_2021} proposes $k$-nearest neighbor MT by searching for target tokens that have similar contextualized representations at each decoding step, an approach further elaborated by \citet{zheng-etal-2021-adaptive,meng-etal-2022-fast} and extended to chunks by \citet{martins-etal-2022-chunkbased}. \fyDone{Check more recent papers ?}%  decide the number of neighbors to retrieve at each step.

% Works using multiple sequences TM: \citep{gu-etal-2018-search}
% % ---
% \paragraph{NAT?}
% 
% NAR + LevT: \citep{gu-etal-2018-nonautoregressive,xiao_survey_2022,xu-carpuat-2021-editor}
% KD: \citep{kim-rush-2016-sequence}
% EDITOR (LevT with swap): \citep{xu-carpuat-2021-editor}
% Lexically-constraint LevT: \citep{susanto-etal-2020-lexically}
% MS-LevT (APE): \citep{kondo-komachi-2022-tmu}

%\paragraph{Retrieval?}
%
% Fuzzy-match repair: \citep{ortega-etal-2016-fuzzy}
% \citep{zhang-etal-2018-guiding} \citep{bulte-tezcan-2019-neural} \citep{xu_boosting_2020}
% \citep{cheng-etal-2022-neural}
% Fuzzy-match prefilter Okapi BM-25 (Apache Lucene): \citep{manning-etal-2019-introduction}
% Cross-lingual external retrieval: \citep{cai-etal-2019-retrieval}
% \paragraph{Other:}
% Multi-align of sequences book: \citep{gusfield-dan-1997-algorithms}
% TM: better productivity \citep{yamada-masaru-2011-effect}
% RATM premise: \citep{feng-etal-2017-memory}
% CAT: \citep{bloodgood-strauss-2014-translation}

% Reuse previous translations NMT (Example Guided NMT): \citep{cao-etal-2019-learning}
%--> 1 src encoder + 1 example encoder
%
% APE vs TM with fuzzy: \citep{sanchez-gijon_post-editing_2019}
% Domain adaptation: \citep{chu-wang-2018-survey}
% $\cdots$

\section{Conclusion and Outlook \label{sec:conclusions}}

In this work, we have extended the Levenshtein Transformer with a new combination operation, making it able to simultaneously edit multiple fuzzy matches and merge them into an initial translation that is then refined. % in the context Example-Based Machine Translation (EBMT)
% , easily applicable for Automatic Post Editing (APE).
Owing to multiple algorithmic contributions and improved training schemes, we have been able to (a) increase the number of output tokens that are copied from retrieved examples; (b) obtain performance improvements compared to using one single match. We have also argued that retrieval-based NMT was a simple way to make the process more transparent for end users.

Next, we would like to work on the retrieval side of the model: first, to increase the diversity of fuzzy matches e.g.\ thanks to contrastive retrieval, but also to study ways to train the retrieval mechanism and extend this approach to search monolingual (target side) corpora. Another line of work will combine our techniques with other approaches to TM-based NMT, such as keeping track of the initial translation(s) on the encoder side.
% Using a neural \textit{fuzzy repair} approach to retrieve the examples, we show that in most domains using several examples can be beneficial even though \textit{fuzzy repair} is not prone to diversity in the retrieved example.

% Our model could benefit from other retrieval techniques. In future works, we would like to design new retrieval methods that could fetch sequences in a large in-domain monolingual dataset and aim for more diversity while not introducing too much noise (due to lower initial modified precision).

\section{Limitations \label{sec:limitations}}

As this work was primarily designed a feasibility study, we have left aside several issues related to performance, which may explain the remaining gap with published results on similar datasets. First, we have restricted the encoder to only encode the source sentence, even though enriching the input side with the initial target(s) has often been found to increase performance \citep{bulte-tezcan-2019-neural}, also for NAT systems \cite{xu-etal-2023-integrating}. It is also likely that increasing the number of training epochs would yield higher absolute scores (see Appendix~\ref{sec:appendix-results}).

These choices were made for the sake of efficiency, as our training already had to fare with the extra computing costs incurred by the alignment procedure required to learn the expert policy. Note that in comparison, the extra cost of the realignment procedure is much smaller, as it is only paid during inference and can be parallelized on GPUs.

We would also like to outline that our systems do not match the performance of an equivalent AR decoder,\fyDone{Add score} a gap that remains for many NAT systems \citep{xiao-etal-2022-survey}. Finally, we have only reported here results for one language pair -- favoring here domain diversity over language diversity -- and would need to confirm the observed improvements on other language pairs and conditions.

% decided to keep our configuration simple
% , it is likely that our NAT decoder is not as efficient as an equivalent AR decoder, as this performance gap has been observed in many studies. 

% For memory reasons, we decided not to include the examples in the encoder, unlike most previous works dealing with \textit{fuzzy repair}. This explains why our method yields lower results than other works with similar setups.

% Some of our data analysis showed that one can increase target coverage up to 15\% using $N$=3 instead of $N$=1 for average matching scores ($\sim 0.4$), but this is a theoretical limit, and the model still suffers from the drawbacks of NAT methods, mainly the multimodality \citep{zhang-etal-2022-study}, and is still not able to fully exploit the examples.

% The new alignment algorithm during training is costly since it has to run on CPU while tensors are stored on GPU. As a matter of fact, it is impossible to implement a DP algorithm without using for loops due to its autoregressive nature.
% Our realignment algorithm implies extra inference time. But this one runs on GPU, therefore being scalable with little delay.

% Most importantly, our new model intends to exploit multiple examples instead of one, but the usual \textit{greedy} retrieval technique does not favor diversity.

% NO ACKNOWLEDGEMENT for the Submission
\section{Acknowledgements}
This work was performed using HPC resources from GENCI-IDRIS (Grant 2022-AD011013583) and Lab-IA from Saclay-IA.

The authors wish to thank Dr. Jitao Xu and Dr. Caio Corro for their guidance and help.
% + the reviewers

% \section{Ethic Statements \label{sec:ethic}}

\bibliography{anthology,custom}

\begin{thebibliography}{44}
\expandafter\ifx\csname natexlab\endcsname\relax\def\natexlab#1{#1}\fi

\bibitem[{Bowker(2002)}]{bowker-2002-computer}
Lynne Bowker. 2002.
\newblock \emph{Computer-aided translation technology: A practical introduction}.
\newblock University of Ottawa Press.

\bibitem[{Bulte and Tezcan(2019)}]{bulte-tezcan-2019-neural}
Bram Bulte and Arda Tezcan. 2019.
\newblock \href {https://doi.org/10.18653/v1/P19-1175} {Neural fuzzy repair: Integrating fuzzy matches into neural machine translation}.
\newblock In \emph{Proceedings of the 57th Annual Meeting of the Association for Computational Linguistics}, pages 1800--1809, Florence, Italy. Association for Computational Linguistics.

\bibitem[{Cai et~al.(2021)Cai, Wang, Li, Lam, and Liu}]{cai-etal-2021-neural}
Deng Cai, Yan Wang, Huayang Li, Wai Lam, and Lemao Liu. 2021.
\newblock \href {https://doi.org/10.18653/v1/2021.acl-long.567} {Neural machine translation with monolingual translation memory}.
\newblock In \emph{Proceedings of the 59th Annual Meeting of the Association for Computational Linguistics and the 11th International Joint Conference on Natural Language Processing (Volume 1: Long Papers)}, pages 7307--7318, Online. Association for Computational Linguistics.

\bibitem[{Carl et~al.(2004)Carl, Way, and Daelemans}]{carl-etal-2004-recent}
Michael Carl, Andy Way, and Walter Daelemans. 2004.
\newblock \href {https://doi.org/10.1162/0891201042544866} {Recent advances in example-based machine translation}.
\newblock \emph{Computational Linguistics}, 30:516--520.

\bibitem[{Carrillo and Lipman(1988)}]{carrillo-etal-1988-multiple}
Humberto Carrillo and David Lipman. 1988.
\newblock \href {https://doi.org/10.1137/0148063} {The multiple sequence alignment problem in biology}.
\newblock \emph{SIAM Journal on Applied Mathematics}, 48(5):1073--1082.

\bibitem[{Cheng et~al.(2022)Cheng, Gao, Liu, Zhao, and Yan}]{cheng-etal-2022-neural}
Xin Cheng, Shen Gao, Lemao Liu, Dongyan Zhao, and Rui Yan. 2022.
\newblock \href {https://doi.org/10.48550/ARXIV.2212.03140} {Neural machine translation with contrastive translation memories}.
\newblock In \emph{Proceedings of the 2022 {Conference} on {Empirical} {Methods} in {Natural} {Language} {Processing} ({EMNLP})}. Association for Computational Linguistics.

\bibitem[{Daum{\'e} et~al.(2009)Daum{\'e}, Langford, and Marcu}]{daume-etal-2009-search}
Hal Daum{\'e}, John Langford, and Daniel Marcu. 2009.
\newblock \href {https://doi.org/10.1007/s10994-009-5106-x} {Search-based structured prediction}.
\newblock \emph{Machine Learning}, 75(3):297--325.

\bibitem[{Garey and Johnson(1979)}]{Garey1979computers}
Michael~R. Garey and David~S. Johnson. 1979.
\newblock Computers and intractability: a guide to the theory of {NP-completeness}.
\newblock W.H. Freeman and Company, New York.

\bibitem[{Gu et~al.(2019)Gu, Wang, and Zhao}]{gu-etal-2019-levenshtein}
Jiatao Gu, Changhan Wang, and Junbo Zhao. 2019.
\newblock \href {https://proceedings.neurips.cc/paper/2019/file/675f9820626f5bc0afb47b57890b466e-Paper.pdf} {Levenshtein transformer}.
\newblock In \emph{Advances in Neural Information Processing Systems}, volume~32. Curran Associates, Inc.

\bibitem[{Gu et~al.(2018)Gu, Wang, Cho, and Li}]{gu-etal-2018-search}
Jiatao Gu, Yong Wang, Kyunghyun Cho, and Victor~O.K. Li. 2018.
\newblock \href {https://doi.org/10.1609/aaai.v32i1.12013} {Search {Engine} {Guided} {Neural} {Machine} {Translation}}.
\newblock \emph{Proceedings of the AAAI Conference on Artificial Intelligence}, 32(1).

\bibitem[{Gusfield(1997)}]{gusfield-dan-1997-algorithms}
Dan Gusfield. 1997.
\newblock \href {https://doi.org/10.1017/CBO9780511574931} {\emph{Algorithms on Strings, Trees, and Sequences: Computer Science and Computational Biology}}.
\newblock Cambridge University Press.

\bibitem[{Guu et~al.(2018)Guu, Hashimoto, Oren, and Liang}]{guu-etal-2018-generating}
Kelvin Guu, Tatsunori~B. Hashimoto, Yonatan Oren, and Percy Liang. 2018.
\newblock \href {https://doi.org/10.1162/tacl_a_00030} {Generating sentences by editing prototypes}.
\newblock \emph{Transactions of the Association for Computational Linguistics}, 6:437--450.

\bibitem[{He et~al.(2021{\natexlab{a}})He, Neubig, and Berg-Kirkpatrick}]{he-etal-2021-efficient}
Junxian He, Graham Neubig, and Taylor Berg-Kirkpatrick. 2021{\natexlab{a}}.
\newblock \href {https://doi.org/10.18653/v1/2021.emnlp-main.461} {Efficient nearest neighbor language models}.
\newblock In \emph{Proceedings of the 2021 Conference on Empirical Methods in Natural Language Processing}, pages 5703--5714, Online and Punta Cana, Dominican Republic. Association for Computational Linguistics.

\bibitem[{He et~al.(2021{\natexlab{b}})He, Huang, Cui, Li, and Liu}]{he-etal-2021-fast}
Qiuxiang He, Guoping Huang, Qu~Cui, Li~Li, and Lemao Liu. 2021{\natexlab{b}}.
\newblock \href {https://doi.org/10.18653/v1/2021.acl-long.246} {Fast and accurate neural machine translation with translation memory}.
\newblock In \emph{Proceedings of the 59th Annual Meeting of the Association for Computational Linguistics and the 11th International Joint Conference on Natural Language Processing (Volume 1: Long Papers)}, pages 3170--3180, Online. Association for Computational Linguistics.

\bibitem[{Hoang et~al.(2022)Hoang, Sachan, Mathur, Thompson, and Federico}]{hoang-etal-2022-improving}
Cuong Hoang, Devendra Sachan, Prashant Mathur, Brian Thompson, and Marcello Federico. 2022.
\newblock \href {https://doi.org/10.48550/arXiv.2210.05047} {Improving {Retrieval} {Augmented} {Neural} {Machine} {Translation} by {Controlling} {Source} and {Fuzzy}-{Match} {Interactions}}.
\newblock ArXiv:2210.05047 [cs].

\bibitem[{Khandelwal et~al.(2021)Khandelwal, Fan, Jurafsky, Zettlemoyer, and Lewis}]{khandelwal_nearest_2021}
Urvashi Khandelwal, Angela Fan, Dan Jurafsky, Luke Zettlemoyer, and Mike Lewis. 2021.
\newblock \href {https://openreview.net/forum?id=7wCBOfJ8hJM} {{Nearest Neighbor Machine Translation}}.
\newblock In \emph{Proceedings of the International Conference on Learning Representations}.

\bibitem[{Kim and Rush(2016)}]{kim-rush-2016-sequence}
Yoon Kim and Alexander~M. Rush. 2016.
\newblock \href {https://doi.org/10.18653/v1/D16-1139} {Sequence-level knowledge distillation}.
\newblock In \emph{Proceedings of the 2016 Conference on Empirical Methods in Natural Language Processing}, pages 1317--1327, Austin, Texas. Association for Computational Linguistics.

\bibitem[{Klein et~al.(2017)Klein, Kim, Deng, Senellart, and Rush}]{klein-etal-2017-opennmt}
Guillaume Klein, Yoon Kim, Yuntian Deng, Jean Senellart, and Alexander Rush. 2017.
\newblock \href {https://aclanthology.org/P17-4012} {{O}pen{NMT}: Open-source toolkit for neural machine translation}.
\newblock In \emph{Proceedings of {ACL} 2017, System Demonstrations}, pages 67--72, Vancouver, Canada. Association for Computational Linguistics.

\bibitem[{Kudo(2018)}]{kudo-2018-subword}
Taku Kudo. 2018.
\newblock \href {https://doi.org/10.18653/v1/P18-1007} {Subword regularization: Improving neural network translation models with multiple subword candidates}.
\newblock In \emph{Proceedings of the 56th Annual Meeting of the Association for Computational Linguistics (Volume 1: Long Papers)}, pages 66--75, Melbourne, Australia. Association for Computational Linguistics.

\bibitem[{Martin et~al.(2020)Martin, Muller, Ortiz~Su{\'a}rez, Dupont, Romary, de~la Clergerie, Seddah, and Sagot}]{martin-etal-2020-camembert}
Louis Martin, Benjamin Muller, Pedro~Javier Ortiz~Su{\'a}rez, Yoann Dupont, Laurent Romary, {\'E}ric de~la Clergerie, Djam{\'e} Seddah, and Beno{\^\i}t Sagot. 2020.
\newblock \href {https://doi.org/10.18653/v1/2020.acl-main.645} {{C}amem{BERT}: a tasty {F}rench language model}.
\newblock In \emph{Proceedings of the 58th Annual Meeting of the Association for Computational Linguistics}, pages 7203--7219, Online. Association for Computational Linguistics.

\bibitem[{Martins et~al.(2022)Martins, Marinho, and Martins}]{martins-etal-2022-chunkbased}
Pedro~Henrique Martins, Zita Marinho, and André F.~T. Martins. 2022.
\newblock \href {https://doi.org/https://aclanthology.org/emnlp-22/2022.emnlp-main.284} {Chunk-based nearest neighbor machine translation}.
\newblock In \emph{Proceedings of the 2022 Conference on Empirical Methods in Natural Language Processing}, pages 4228--4245, Abu Dhabi, United Arab Emirates. Association for Computational Linguistics.

\bibitem[{Meng et~al.(2022)Meng, Li, Zheng, Wu, Sun, Zhang, and Li}]{meng-etal-2022-fast}
Yuxian Meng, Xiaoya Li, Xiayu Zheng, Fei Wu, Xiaofei Sun, Tianwei Zhang, and Jiwei Li. 2022.
\newblock \href {https://doi.org/10.18653/v1/2022.findings-acl.47} {Fast nearest neighbor machine translation}.
\newblock In \emph{Findings of the Association for Computational Linguistics: ACL 2022}, pages 555--565, Dublin, Ireland. Association for Computational Linguistics.

\bibitem[{Nagao(1984)}]{nagao-1984-framework}
Makoto Nagao. 1984.
\newblock A framework of a mechanical translation between {Japanese} and {English} by analogy principle.
\newblock In \emph{Artificial and human intelligence}. Elsevier Science Publishers. B.V.

\bibitem[{Niwa et~al.(2022)Niwa, Takase, and Okazaki}]{niwa-etal-2022-nearest}
Ayana Niwa, Sho Takase, and Naoaki Okazaki. 2022.
\newblock \href {https://doi.org/10.48550/ARXIV.2208.12496} {Nearest neighbor non-autoregressive text generation}.
\newblock \emph{CoRR}, abs/2208.12496.

\bibitem[{Papineni et~al.(2002)Papineni, Roukos, Ward, and Zhu}]{papineni-etal-2002-bleu}
Kishore Papineni, Salim Roukos, Todd Ward, and Wei-Jing Zhu. 2002.
\newblock \href {https://doi.org/10.3115/1073083.1073135} {{B}leu: a method for automatic evaluation of machine translation}.
\newblock In \emph{Proceedings of the 40th Annual Meeting of the Association for Computational Linguistics}, pages 311--318, Philadelphia, Pennsylvania, USA. Association for Computational Linguistics.

\bibitem[{Pham et~al.(2020)Pham, Xu, Crego, Yvon, and Senellart}]{pham-etal-2020-priming}
Minh~Quang Pham, Jitao Xu, Josep Crego, Fran{\c{c}}ois Yvon, and Jean Senellart. 2020.
\newblock \href {https://aclanthology.org/2020.wmt-1.63} {Priming neural machine translation}.
\newblock In \emph{Proceedings of the Fifth Conference on Machine Translation}, pages 516--527, Online. Association for Computational Linguistics.

\bibitem[{Popovi{\'c}(2015)}]{popovic-2015-chrf}
Maja Popovi{\'c}. 2015.
\newblock \href {https://doi.org/10.18653/v1/W15-3049} {chr{F}: character n-gram {F}-score for automatic {MT} evaluation}.
\newblock In \emph{Proceedings of the Tenth Workshop on Statistical Machine Translation}, pages 392--395, Lisbon, Portugal. Association for Computational Linguistics.

\bibitem[{Post(2018)}]{post-2018-call}
Matt Post. 2018.
\newblock \href {https://doi.org/10.18653/v1/W18-6319} {A call for clarity in reporting {BLEU} scores}.
\newblock In \emph{Proceedings of the Third Conference on Machine Translation: Research Papers}, pages 186--191, Brussels, Belgium. Association for Computational Linguistics.

\bibitem[{Rei et~al.(2020)Rei, Stewart, Farinha, and Lavie}]{rei-etal-2020-comet}
Ricardo Rei, Craig Stewart, Ana~C Farinha, and Alon Lavie. 2020.
\newblock \href {https://doi.org/10.18653/v1/2020.emnlp-main.213} {{COMET}: A neural framework for {MT} evaluation}.
\newblock In \emph{Proceedings of the 2020 Conference on Empirical Methods in Natural Language Processing (EMNLP)}, pages 2685--2702, Online. Association for Computational Linguistics.

\bibitem[{Ross et~al.(2011)Ross, Gordon, and Bagnell}]{ross-etal-2011-reduction}
Stephane Ross, Geoffrey Gordon, and Drew Bagnell. 2011.
\newblock \href {https://proceedings.mlr.press/v15/ross11a.html} {A reduction of imitation learning and structured prediction to no-regret online learning}.
\newblock In \emph{Proceedings of the Fourteenth International Conference on Artificial Intelligence and Statistics}, volume~15 of \emph{Proceedings of Machine Learning Research}, pages 627--635, Fort Lauderdale, FL, USA. PMLR.

\bibitem[{Rudin(2019)}]{rudin-cynthia-2019-stop}
Cynthia Rudin. 2019.
\newblock \href {https://doi.org/10.1038/s42256-019-0048-x} {Stop explaining black box machine learning models for high stakes decisions and use interpretable models instead}.
\newblock \emph{Nature Machine Intelligence}, 1(5):206--215.

\bibitem[{Somers(1999)}]{somers-1999-review}
Harold Somers. 1999.
\newblock \href {https://doi.org/10.1023/A:1008109312730} {Review article: Example-based machine translation}.
\newblock \emph{Machine Translation}, 14(2):113--157.

\bibitem[{Stern et~al.(2019)Stern, Chan, Kiros, and Uszkoreit}]{stern-etal-2019-insertion}
Mitchell Stern, William Chan, Jamie Kiros, and Jakob Uszkoreit. 2019.
\newblock \href {https://proceedings.mlr.press/v97/stern19a.html} {Insertion transformer: Flexible sequence generation via insertion operations}.
\newblock In \emph{Proceedings of the 36th International Conference on Machine Learning}, volume~97 of \emph{Proceedings of Machine Learning Research}, pages 5976--5985. PMLR.

\bibitem[{Susanto et~al.(2020)Susanto, Chollampatt, and Tan}]{susanto-etal-2020-lexically}
Raymond~Hendy Susanto, Shamil Chollampatt, and Liling Tan. 2020.
\newblock \href {https://doi.org/10.18653/v1/2020.acl-main.325} {Lexically constrained neural machine translation with {L}evenshtein transformer}.
\newblock In \emph{Proceedings of the 58th Annual Meeting of the Association for Computational Linguistics}, pages 3536--3543, Online. Association for Computational Linguistics.

\bibitem[{Vaswani et~al.(2017)Vaswani, Shazeer, Parmar, Uszkoreit, Jones, Gomez, Kaiser, and Polosukhin}]{vaswani-etal-2017-attention}
Ashish Vaswani, Noam Shazeer, Niki Parmar, Jakob Uszkoreit, Llion Jones, Aidan~N. Gomez, \L{}ukasz Kaiser, and Illia Polosukhin. 2017.
\newblock Attention is all you need.
\newblock In \emph{Proceedings of the 31st International Conference on Neural Information Processing Systems}, NIPS'17, page 6000–6010, Red Hook, NY, USA. Curran Associates Inc.

\bibitem[{Xia et~al.(2019)Xia, Huang, Liu, and Shi}]{xia-etal-2019-graph}
Mengzhou Xia, Guoping Huang, Lemao Liu, and Shuming Shi. 2019.
\newblock \href {https://doi.org/10.1609/aaai.v33i01.33017297} {Graph based translation memory for neural machine translation}.
\newblock \emph{Proceedings of the AAAI Conference on Artificial Intelligence}, 33(01):7297--7304.

\bibitem[{Xiao et~al.(2023)Xiao, Wu, Guo, Li, Zhang, Qin, and Liu}]{xiao-etal-2022-survey}
Yisheng Xiao, Lijun Wu, Junliang Guo, Juntao Li, Min Zhang, Tao Qin, and Tie-Yan Liu. 2023.
\newblock \href {https://doi.org/10.1109/TPAMI.2023.3277122} {A survey on non-autoregressive generation for neural machine translation and beyond}.
\newblock \emph{IEEE Transactions on Pattern Analysis and Machine Intelligence}, pages 1--20.

\bibitem[{Xu et~al.(2020)Xu, Crego, and Senellart}]{xu-etal-2020-boosting}
Jitao Xu, Josep Crego, and Jean Senellart. 2020.
\newblock \href {https://doi.org/10.18653/v1/2020.acl-main.144} {Boosting neural machine translation with similar translations}.
\newblock In \emph{Proceedings of the 58th Annual Meeting of the Association for Computational Linguistics}, pages 1580--1590, Online. Association for Computational Linguistics.

\bibitem[{Xu et~al.(2023)Xu, Crego, and Yvon}]{xu-etal-2023-integrating}
Jitao Xu, Josep Crego, and Fran{\c{c}}ois Yvon. 2023.
\newblock \href {https://aclanthology.org/2023.eacl-main.96} {Integrating translation memories into non-autoregressive machine translation}.
\newblock In \emph{Proceedings of the 17th Conference of the European Chapter of the Association for Computational Linguistics}, pages 1326--1338, Dubrovnik, Croatia. Association for Computational Linguistics.

\bibitem[{Xu and Carpuat(2021)}]{xu-carpuat-2021-editor}
Weijia Xu and Marine Carpuat. 2021.
\newblock \href {https://doi.org/10.1162/tacl_a_00368} {{EDITOR}: An edit-based transformer with repositioning for neural machine translation with soft lexical constraints}.
\newblock \emph{Transactions of the Association for Computational Linguistics}, 9:311--328.

\bibitem[{Zhang et~al.(2018)Zhang, Utiyama, Sumita, Neubig, and Nakamura}]{zhang-etal-2018-guiding}
Jingyi Zhang, Masao Utiyama, Eiichro Sumita, Graham Neubig, and Satoshi Nakamura. 2018.
\newblock \href {https://doi.org/10.18653/v1/N18-1120} {Guiding neural machine translation with retrieved translation pieces}.
\newblock In \emph{Proceedings of the 2018 Conference of the North {A}merican Chapter of the Association for Computational Linguistics: Human Language Technologies, Volume 1 (Long Papers)}, pages 1325--1335, New Orleans, Louisiana. Association for Computational Linguistics.

\bibitem[{Zheng et~al.(2023)Zheng, Wang, Wang, Chen, Zhang, and Tu}]{zheng-etal-2023-towards}
Kangjie Zheng, Longyue Wang, Zhihao Wang, Binqi Chen, Ming Zhang, and Zhaopeng Tu. 2023.
\newblock \href {https://doi.org/10.1109/ICASSP49357.2023.10094646} {Towards a unified training for {Levenshtein} transformer}.
\newblock In \emph{Proceedings of the IEEE International Conference on Acoustics, Speech and Signal Processing (ICASSP)}, pages 1--5.

\bibitem[{Zheng et~al.(2021)Zheng, Zhang, Guo, Huang, Chen, Luo, and Chen}]{zheng-etal-2021-adaptive}
Xin Zheng, Zhirui Zhang, Junliang Guo, Shujian Huang, Boxing Chen, Weihua Luo, and Jiajun Chen. 2021.
\newblock \href {https://doi.org/10.18653/v1/2021.acl-short.47} {Adaptive nearest neighbor machine translation}.
\newblock In \emph{Proceedings of the 59th Annual Meeting of the Association for Computational Linguistics and the 11th International Joint Conference on Natural Language Processing (Volume 2: Short Papers)}, pages 368--374, Online. Association for Computational Linguistics.

\bibitem[{Zhou et~al.(2020)Zhou, Gu, and Neubig}]{zhou-2021-understanding}
Chunting Zhou, Jiatao Gu, and Graham Neubig. 2020.
\newblock \href {https://openreview.net/forum?id=BygFVAEKDH} {Understanding knowledge distillation in non-autoregressive machine translation}.
\newblock In \emph{Proceedings of the International Conference on Learning Representations}.

\end{thebibliography}
\bibliographystyle{acl_natbib}

\appendix

\section{Model Configuration}
\label{sec:appendix-config}

We use a Transformer architecture with embeddings of dimension $512$; feed-forward layers of size $2048$; number of heads $8$; number of encoder and decoder layers: 6; batch size: 3000 tokens; shared-embeddings; dropout: 0.3; number of GPUs: 6. The maximal number of additional placeholders is $K_{max}=64$.
%
% Encoder weights are initialized with BERT \citep{devlin-etal-2019-bert}.\fyTodo{Does this matter?}

During training, we use Adam optimizer with ($\beta_1$, $\beta_2$)=(0.9, 0.98); inverse sqrt scheduler; learning rate: $5e^{-4}$; label smoothing: 0.1; warmup updates: 10,000; float precision: 16. We fixed the number of iterations at $60$k.
For decoding, we use iterative refinement with an empty placeholder penalty of $3$, and a max number of iterations of $10$ \cite{gu-etal-2019-levenshtein}.

For the $n$-way alignment (\textsection{}\ref{ssec:optimal-alignment:formulation}), we use $k$=$10$.

The hyper-parameters of the realigner (\textsection\ref{sec:appendix-realignment}) were tuned on a subset of 1k samples extracted from the ECB training set.

Metrics are used with default settings: SacreBLEU signature is
\texttt{nrefs:1|case:mixed|eff:no} \texttt{|tok:13a|smooth:exp|version:2.1.0};
the ChrF signature is \texttt{nrefs:1|case:mixed|eff:yes} \texttt{|nc:6|nw:0|space:no|version:2.1.0};
as for COMET we use the default model of version 1.1.3: \texttt{Unbabel/wmt22-comet-da}.
\fyDone{Give version and parameters}

\section{Data Analysis}
\label{sec:appendix-data}

Table~\ref{table:data-stats} contains statistics about all 11 domains. They notably highlight the relationship between the average number of retrieved sentences during training and the ability of \mlevt{3} to perform better than \mlevt{1} in Table~\ref{table:BLEU-num-example}. The domains with retrieval rates lesser than $1$ (Epp, News, TED, Ubu)
have quite a broad content,
% are quite broad in content, 
meaning that training instances have fewer close matches, which also means that for these domains, \mlevt{3} hardly sees two or three examples that it needs to use in inference.
\begin{table*}[ht]
	\centering
	\scalebox{0.9}{
		\begin{tabular}{lrrrrrrrrrrrr}
			\hline
			\rowcolor{LightGray}
			domain         & ECB  & EME  & Epp  & GNO  & JRC  & KDE  & News & PHP  & TED  & Ubu  & Wiki & all  \\ \hline
			size           & 195k & 373k & 2.0M & 55k  & 503k & 180k & 151k & 16k  & 159k & 9k   & 803k & 4.4M \\ \hdashline
			retrieval rate & 1.66 & 2.12 & 0.91 & 1.18 & 1.71 & 1.10 & 0.24 & 1.20 & 0.96 & 0.62 & 1.22 & 1.18 \\ \hdashline
			mean length    & 29.2 & 16.7 & 26.6 & 9.4  & 28.8 & 10.5 & 26.4 & 14.5 & 17.7 & 5.2  & 19.6 & 18.6 \\
		\end{tabular}
	}
	\caption{\label{table:data-stats}Number of samples, average number of retrieved sentences and average length of sentences after tokenization for all 11 domains.}
\end{table*}

\section{Realignment}
\label{sec:appendix-realignment}
The realignment process is an extra inference step aiming to improve the result of the placeholder insertion stage.
% We are in the situation when our multi-lev model, at inference, during the first pass, predicts placeholder insertions \verb|<PLH>|.
To motivate our approach, let us consider the following sentences before placeholder insertion:

\begin{tabular}{ccccccc}
	$\trg_0^{plh}$: & $<$ & A & B & C   & $>$      & $\times$ \\
	$\trg_1^{plh}$: & $<$ & B & C & $>$ & $\times$ & $\times$ \\
	$\trg_2^{plh}$: & $<$ & A & D & C   & D        & $>$,
\end{tabular}\\
where letters represent tokens, $\times$ denotes padding, $<$ and $>$ respectively stand for \verb|<BOS>| and \verb|<EOS>|.

The output of this stage is a prediction for all pairs of consecutive tokens. This prediction takes the form of a tensor $\log \pi_\theta^{plh}$ of dimensions $N \times (L - 1) \times (K_{max}+1)$, corresponding respectively to the number of retrieved sentences $N$, the maximum sentence length $L$, and the maximum number of additional placeholders $K_{max}$.

Let $P$ (a $N \times (L-1)$ tensor) denote the $\argmax$, e.g.\:
$P = $ \begin{tabular}{ccccccc}
	0 & 0 & 0 & 2 & 0 \\
	0 & 0 & 1 & 0 & 0 \\
	0 & 1 & 0 & 0 & 0
\end{tabular}

Inserting the prescribed number of placeholders (figured by $\_$) then yields the following $\trg^{cmb}$:
\begin{tabular}{ccccccccc}
	$\trg_0^{cmb}$: & $<$ & A & B  & C  & \_  & \_       & $>$      \\
	$\trg_1^{cmb}$: & $<$ & B & C  & \_ & $>$ & $\times$ & $\times$ \\
	$\trg_2^{cmb}$: & $<$ & A & \_ & D  & C   & D        & $>$
\end{tabular}

This result is far from perfect, as it fails to align the repeated occurrences of $C$. For instance, a preferable alignment requiring $3$ changes ($1$ change consists in a modification of $\pm 1$ in $P$) could be:

\begin{tabular}{ccccccccc}
	${\trg_0^{cmb}}'$: & $<$ & A  & B & C & \_ & $>$ & $\times$ \\
	${\trg_1^{cmb}}'$: & $<$ & \_ & B & C & \_ & $>$ & $\times$ \\
	${\trg_2^{cmb}}'$: & $<$ & A  & D & C & D  & $>$ & $\times$
\end{tabular}

The general goal of realignment is to improve such alignments by performing a small number of changes in $P$.
We formalize this problem as a search for a good tradeoff between (a) the individual placeholder prediction scores, aggregated in $\mathcal L_L$ (likelihood loss) and (b) $\mathcal L_A$ an alignment loss. Under its simplest form, this problem is again an optimal multisequence alignment problem, for which exact dynamic programming solutions are computationally intractable in our setting.

We instead develop a continuous relaxation that can be solved effectively with SGD and is also easy to parallelize on GPUs. We, therefore, relax the integer condition for $P$ and assume that $P_{i,j}$ can take continuous real values in $[0, K_{max}]$, then solve the continuous optimization problem before turning $P_{i,j}$ values back into integers.
% We keep calling $P$ the continuous version matrix.
% This technique may still have the drawback of being unrobust to the rounding a shift the sequences and provide still a misalignment... But we will come back to that last point with a fix

The likelihood loss aims to keep the $P_{i,j}$ values close to the model predictions. Denoting $(\mu, \sigma)$ respectively the mean and variance of the model predictions, our initial version of this loss is
\[
	\mathcal L_L(P) = \sum_{i, j} \frac{(P_{i, j} - \mu_{i,j})^2}{2 \sigma^2}.
\]
In practice, we found that using a weighted average $\hat \mu$ and clamping the variance $\hat \sigma^2$ both yield better realignments, yielding:

\[
	\mathcal L_L(P) = \sum_{i, j} \frac{(P_{i, j} - \hat \mu_{i,j})^2}{2 \hat \sigma^2}
\]
% We assumed that the logit distribution is a discretization to a categorical distribution from a normal distribution $\mathcal N (\mu, \sigma)$.
% Then $\mu$ can be estimated as the expectation of the logit distribution, and $\sigma^2$ as the variance.
% In practice, 
% If we define $p_k$ as the softmax of a given logit distribution $x_0, \cdots, x_{K_{max}}$:
% \[
% 	p_k = \frac{e^{x_k}}{\sum_l e^{x_l}}
% \]

% Then the mean and the variance of the normal distribution are estimated as the ones of the categorical distribution parametrized by $p_k$:
% \begin{align*}
% 	\mu &= \sum_k k p_k \\
% 	\sigma^2 &= \sum_k k^2 p_k - \mu^2
% \end{align*}

% The new discrete distribution is then:
% \[
% 	p(t) = \frac{1}{Z} e^{-\frac{(t - \mu)^2}{2 \sigma^2}}
% \]

% Since the \textit{logit function} is the log of the probabilities:

% \[
% 	l(t) = -\frac{(t - \mu)^2}{2 \sigma^2} - \log Z
% \]
% And since only the derivative really matters, we can remove the constant term $\log Z$.

% The \textit{logit function} $l_{i j}$ is eventually:
% \[
% 	l_{ij}(p) = -\frac{(t - \mu_{ij})^2}{2 \sigma_{ij}^2}
% \]

% Our \textit{logit loss} is finally defined as:

% \[
% 	\mathcal L_L(P) = -\sum_{i = 0}^{N-1} \sum_{j = 0}^{L-2} l_{ij}(P_{i,j})
% \]

To define the \textit{alignment loss}, we introduce a \emph{position matrix} $X$ of dimension $N \times L$ in $\mathbb R^+$, where $X_{n, i}$ corresponds to the (continuous)position of token $y_{n,i}$ after inserting a real number of placeholders. $X$ is defined as:
% \fyTodo{On passe x à y pour les séquences sans raison.}
% C'est pcq les y sont les séquences de tokens, et X les positions.

\[
	X_{n, i}(P) = i + \sum_{j < i} P_{n, j}
\]

with $i$ the number of tokens occuring before $X_{n, i}$ and $\sum_{j < i} P_{n, j}$ the cumulated sum of placeholders. Using $X$, we derive the distance tensor $D$ of dimension $N \times L \times N \times L$ in $\mathbb R^+$ as:
\[
	D_{n,i,m,j}(P) = |X_{n, i} - X_{m, j}|
\]

Finally, let $G$ be an $N \times L \times N \times L$ alignment graph tensor, where $G_{n, i, m, j} = 1$ if and only if $y_{n, i} = y_{m, j}$ and $n \neq m$ and $D_{n, i, m, j} < D_{max}$. $G$ connects identical tokens in different sentences when their distance after placeholder insertion is at most $D_{max}$. This last condition avoids perturbations from remote tokens that coincidentally appear to be identical.

Each token $y_{n, i}$ is associated with an individual loss:

\[
	d_{n, i}(P) = \left\{
	\begin{array}{ll}
		\underset{m, j}{\min} \left\{ D_{n, i, m, j}(P) : G_{n, i, m, j} = 1 \right\} \\
		\quad\text{if } \exists (m,j) \text{ s.t. }  G_{n, i, m, j} = 1               \\
		0 \text{~~otherwise.}
	\end{array}
	\right.
\]

The \textit{alignment loss} aggregates these values over sentences and positions as:

\[
	\mathcal L_A(P) = \sum_{n = 0}^{N-1} \sum_{i = 0}^{L-1}  d_{n, i}(P)
\]

A final ingredient in our realignment model is related to the final discretization step. To avoid rounding errors, we further constrain the optimization process to deliver near-integer solutions. For this, we also include a \textit{integer constraint} loss defined as :
\[
	\mathcal{L}_{int}(P) = \mu_t \sum_{i, j} \sin^2 (\pi P_{i, j})
\]
where $\mu_t$ controls the scale of $\mathcal{L}_{int}(P)$. As $x \rightarrow{} \sin^2 (\pi x)$ reaches its minimum 0 for integer values, minimizing $\mathcal{L}_{int}(P)$ has the effect of enforcing a near-integer constraint to our solutions.  Overall, we minimize in $P$:
\[
	\mathcal{L} = \mathcal L_L(P) + \mathcal L_A(P) + \mathcal{L}_{int}(P),
\]
slowly increasing the scale of $\mu_t$ according to the following schedule
\[
	\mu_t = \left\{
	\begin{array}{ll}
		0                                     & \text{ if } t < t_0 \\
		\mu_T                                 & \text{ if } t > T   \\
		\mu_T \frac{(t - t_0)^2}{(T - t_0)^2} & \text{ otherwise}
	\end{array}
	\right.,
\]
with $t_0$, $T$ the timestamps for respectively the activation of the \textit{integer constraint} loss, and the activation of the clamping.
This optimization is performed with gradient descent directly on GPUs, with a small additional cost to the inference procedure.

\section{NP-hardness of Coverage Maximization in N-way Alignment}
\label{sec:appendix:NP-hard}

Given the set of possible N-way alignments, the problem of finding the one that maximizes the target coverage is NP-hard. To prove it, we can reduce the NP-hard \textit{set cover} problem \citep{Garey1979computers} to the \textit{N-way alignment coverage maximization} problem.

\begin{itemize}
	\item \textit{Cover set} decision problem (A):

	      Let $X=\{x_1, \cdots, x_N\}$ and $ C_0 \subset 2^X$. Is there $c^*=(c_1, \cdots, c_K) \in  C_0^K$ s.t. $|\cup_k c_k^*| = |X|$?

	\item \textit{N-way alignment coverage maximization} decision problem (B):

	      Let $X=\{x_1, \cdots, x_N\}$ \\and $C=(C_1, \cdots, C_K) \subset (2^X)^K$.
	      For $p \in \mathbb N$, is there $c \in \prod_{k=1}^K C_k$ s.t. $|\cup_k c_k| \geq p$?
\end{itemize}

%\begin{newtext}
A solution of (B) can be certified in polynomial time: we simply compute the cardinal of a union.
Any instance of (A) can be transformed in polynomial time and space into a special instance of (B) where all $C_k = C_0$ and $p = |X|$.
%\end{newtext}
\fyDone{Dire que le problème est dans P. Donner des éléments de la réduction}

\section{Results for fr-en}
\label{sec:reverse-dir}

Table~\ref{table:benefit-N-reverse-dir} reports the BLEU scores for the reverse direction (fr$\rightarrow$en), using exactly the same configuration as in Table~\ref{table:benefit-N}. Note that since we used the same data split (retrieving examples based on the similarity in English), and since the retrieval procedure is asymmetrical, 4,749 test samples happen to have no match. That would correspond to an extra column labeled "0", which is not represented here.

\begin{table}[h]
	\centering
	\scalebox{1}{
		% \begin{tabular}{lrrrr}
		\begin{tabular}{p{0.09\textwidth}>{\centering}p{0.06\textwidth}>{\centering}p{0.06\textwidth}>{\centering\arraybackslash}p{0.06\textwidth}>{\centering\arraybackslash}p{0.06\textwidth}}
			Model \textbackslash N     & 1     & 2     & 3      & all    \\ \hline
			% \rowcolor{LightGray}
			size                       & 2,753 & 1,675 & 12,823 & 17,251 \\
			\hdashline
			\mlevt{1}                  & 57.2  & 57.6  & 61.5   & 60.2   \\
			\mlevt{3} \texttt{+pt +ra} & 58.2  & 59.4  & 64.0   & 62.4   \\
		\end{tabular}
	}
	\caption{\label{tab:benefit-N-reverse-dir} BLEU scores on the full test set. \mlevt{3} is improved with pre-training and realignment. All BLEU differences are significant ($p=0.05$). $p$-values from SacreBLEU paired bootstrap resampling $(n=1000)$.}
\end{table}

The reverse direction follows a similar pattern, providing further evidence of the method's effectiveness.

\section{Complementary Analyses}
\label{sec:appendix-results}

\paragraph{Diversity and difficulty}
Results in Table~\ref{table:BLEU-num-example} show that some datasets do not seem to benefit from multiple examples. This is notably the case for Europarl, News-Commentary, TED2013, and Ubuntu. We claim that this was due to the lack of retrieved examples at training (as stated in \textsection{}\ref{sec:appendix-data}), of diversity, and the noise in fuzzy matches.
To further investigate this issue, we report two scores in Table~\ref{table:datasets-difficulty}. The first is the increase of bag-of-word coverage of the target gained by using $N$=$3$ instead of $N$=$1$; the second is the increase of noise in the examples, computed as the proportion of tokens in the examples that do not occur in the target. We observe that, in fact, low diversity is often associated with poor scores for \mlevt{3}, and higher diversity with better performance.
%, these scores can explain why some datasets are more "difficult".
% Still, it does not explain in itself the poor performances on TED2013 and Europarl.

\begin{table}[h]
	\centering
	\begin{tabular}{lrrrr}
		     & \multicolumn{2}{c}{cover} & \multicolumn{2}{c}{noise}                                         \\
		     & \textit{test-0.4}         & \textit{test-0.6}         & \textit{test-0.4} & \textit{test-0.6} \\ \hline
		ECB  & +8.2                      & +8.9                      & +3.7              & +4.4              \\
		EME  & +8.9                      & +8.5                      & +4.0              & +5.0              \\
		Epp  & +10.5                     & +13.7                     & +2.2              & +4.0              \\
		GNO  & +\textbf{7.1}             & +\textbf{6.2}             & +\textbf{7.6}     & +\textbf{9.3}     \\
		JRC  & +\textbf{7.2}             & +\textbf{6.8}             & +4.8              & +5.2              \\
		KDE  & +8.0                      & +\textbf{7.5}             & +\textbf{6.9}     & +\textbf{7.8}     \\
		News & +\textbf{7.2}             & +12.5                     & +2.3              & +5.0              \\
		PHP  & +\textbf{7.2}             & +\textbf{7.6}             & +4.2              & +5.4              \\
		TED  & +9.4                      & +11.4                     & +2.9              & +4.6              \\
		Ubu  & +\textbf{5.5}             & +\textbf{6.0}             & +\textbf{6.7}     & +\textbf{8.6}     \\
		Wiki & +8.0                      & +8.0                      & +2.5              & +3.6              \\
		all  & +8.1                      & +8.9                      & +4.4              & +5.7              \\
	\end{tabular}
	\caption{\label{table:datasets-difficulty} Coverage and noise scores increase. "Difficulty" is highlighted in bold ($<8.0$ for cover; $>6.0$ for noise).}
\end{table}

\paragraph{Long run} All results in the main text were obtained with models trained for 60k iterations, which was enough to compare the various models while saving computation resources. For completeness, we also performed one longer training for 300k iterations for \mlevt{3} (see Table~\ref{table:long-run}), which resulted in an improvement of around +2 BLEU for each test set. This is without realignment nor pretraining.
% \fyDone{Also mLevT-1, non ?}\fyDone{Without pretraining?}

\begin{table}[!h]
	\centering
	\begin{tabular}{llrr}
		model     &           & \textit{test-0.4} & \textit{test-0.6} \\ \hline
		\mlevt{3} &           & 46.5              & 60.0              \\
		          & + realign & 46.7              & 60.2              \\
		\mlevt{3} & long      & 48.7              & 61.9              \\
		          & + realign & \textbf{48.9}     & \textbf{62.0}     \\
	\end{tabular}
	\caption{\label{table:long-run} BLEU score of \mlevt{3}: 60k iterations; and \mlevt{3} \texttt{long}: 300k iterations.}
\end{table}

\paragraph{The Benefits of realignment} Table~\ref{table:realignment-num-pass} shows that realignment also decreases the average number of refinement steps to converge. These results suggest that the edition is made easier with realignment.

\begin{table}[h]
	\centering
	\begin{tabular}{llrr}
		model     &          & \textit{test-0.4} & \textit{test-0.6} \\ \hline
		\mlevt{3} &          & 3.55              & 2.07              \\
		          & +realign & \textbf{3.37}     & \textbf{1.93}     \\
	\end{tabular}
	\caption{\label{table:realignment-num-pass} Average number of extra refinement rounds.}
\end{table}

In Table~\ref{table:mod-prec-all}, we present detailed results of the unigram modified precision of \levt, \mlevt{3} and \mlevt{3}\texttt{+realign}. Using more examples indeed increases copy (+4.4), even though it diminishes copy precision (-1.7). Again we observe the positive effect of realignment, which amplifies the tendency of our model to copy input tokens.
% \maxSmallTodo{Again? --> les proportions sont déjà présentées dans les résultats principaux...}

\begin{table}[h]
	\centering
	\begin{tabular}{llrr}
		model      &               & precision     & \% units      \\ \hline
		\tmlevt    & \textit{copy} & \textbf{87.5} & 64.9          \\
		           & \textit{gen}  & 52.6          & 35.1          \\
		\mlevt{3}  & \textit{copy} & 85.4          & 68.8          \\
		           & \textit{gen}  & 54.9          & 31.2          \\
		~~+realign & \textit{copy} & 85.8          & \textbf{69.3} \\
		           & \textit{gen}  & 54.7          & 30.7          \\
	\end{tabular}
	\caption{\label{table:mod-prec-all} Modified precision of copy vs. generated unigrams of \levt{} vs. \mlevt{3}.}
\end{table}

%\begin{newtext}
\paragraph{COMET scores} We compute COMET scores \citep{rei-etal-2020-comet} separately for each domain with default \texttt{wmt20-comet-da} similarly to Table~\ref{table:BLEU-num-example} (see Table~\ref{table:comet-num-example}). We observe that the basic version of \mlevt{3}underperforms \mlevt{1}; we also see a great variation in the scores. A possible explanation can be a fluency decline when using multiple examples, which is not represented by the precision scores computed by BLEU. The improved version, using realignment and pre-training, confirms that adding more matches is overall beneficial for MT quality.
% \end{newtext}
\begin{table*}[ht]
	\centering
	\scalebox{0.9}{
		\begin{tabular}{llrrrrrrrrrrr|r}
			\hline
			\rowcolor{LightGray}
			                  &                                   & ECB           & EME           & Epp           & GNO           & JRC           & KDE           & \makebox[\widthof{ECB}]{News} & PHP            & TED           & Ubu           & \makebox[\widthof{ECB}]{Wiki} & \textbf{all}  \\ \hline
			\textit{test-0.4} & \mlevt{1}                         & 33.0          & 43.1          & 39.9          & 56.3          & 70.7          & 37.4          & -2.0                          & -39.6          & -0.8          & 41.6          & -9.9                          & 24.5          \\[-1pt] \hdashline
			                  & \mlevt{2}                         & 27.2          & 42.0          & 31.1          & 48.0          & 64.4          & 32.6          & -10.8                         & -42.7          & -8.7          & 35.3          & -15.7                         & 18.4          \\[-1pt] \hdashline
			                  & \mlevt{3}                         & 27.3          & 42.1          & 26.8          & 51.5          & 64.2          & 33.5          & -10.1                         & -39.9          & -14.8         & 38.6          & -16.3                         & 18.4          \\[-1pt] \hdashline
			                  & \multicolumn{1}{l}{~~~+pre-train} & 30.6          & \textbf{44.7} & 38.7          & \textbf{57.9} & 67.8          & \textbf{37.9} & -6.2                          & \textbf{-35.3} & \textbf{2.7}  & 41.5          & \textbf{-5.0}                 & \textbf{25.0} \\[-1pt] \hdashline
			                  & \multicolumn{1}{l}{~~~+realign}   & 31.0          & \textbf{45.1} & 32.0          & 53.2          & 66.4          & 34.9          & -10.9                         & -39.0          & -11.9         & 41.1          & -10.8                         & 21.0          \\[-1pt] \hdashline
			                  & \multicolumn{1}{l}{~~~+both}      & \textbf{33.7} & \textbf{46.4} & \textbf{42.4} & \textbf{59.9} & \textbf{69.9} & \textbf{40.1} & \textbf{-1.0}                 & \textbf{-33.0} & \textbf{5.1}  & \textbf{43.5} & \textbf{-3.7}                 & \textbf{27.5} \\[-1pt]
			\hline\hline
			\textit{test-0.6} & \mlevt{1}                         & 51.7          & 53.9          & 56.9          & 65.3          & 85.4          & 37.0          & -3.0                          & -17.0          & 48.5          & 57.6          & 64.5                          & 45.5          \\[-1pt] \hdashline
			                  & \mlevt{2}                         & 51.2          & \textbf{54.2} & 56.3          & 64.2          & 82.5          & 34.6          & -9.0                          & \textbf{-15.9} & 46.1          & 55.5          & 61.6                          & 43.7          \\[-1pt] \hdashline
			                  & \mlevt{3}                         & 50.9          & \textbf{55.8} & 54.1          & 65.1          & 81.1          & 33.6          & -9.8                          & \textbf{-16.2} & 41.1          & 57.4          & 61.7                          & 43.1          \\[-1pt] \hdashline
			                  & \multicolumn{1}{l}{~~~+pre-train} & \textbf{54.6} & \textbf{56.5} & \textbf{58.0} & \textbf{68.2} & 84.7          & \textbf{41.5} & -4.3                          & \textbf{-11.8} & \textbf{48.9} & \textbf{62.7} & \textbf{65.6}                 & \textbf{47.7} \\[-1pt] \hdashline
			                  & \multicolumn{1}{l}{~~~+realign}   & \textbf{53.0} & \textbf{56.4} & 56.3          & 65.4          & 83.6          & 34.3          & -7.2                          & \textbf{-16.1} & 42.8          & \textbf{58.3} & 63.7                          & 44.6          \\[-1pt] \hdashline
			                  & \multicolumn{1}{l}{~~~+both}      & \textbf{55.7} & \textbf{57.4} & \textbf{58.8} & \textbf{70.5} & \textbf{85.8} & \textbf{43.2} & -4.4                          & \textbf{-10.6} & \textbf{49.4} & \textbf{62.3} & \textbf{67.7}                 & \textbf{48.7} \\[-1pt]
		\end{tabular}
	}
	\caption{\label{table:comet-num-example} Per domain COMET scores (x 100) for \mlevt{n} and variants. Bold for scores better than \mlevt{1}.}
	% Italic results are less than 0.5 BLEU under \mlevt{1}.}
\end{table*}

\paragraph{Per-domain ablation study} Table~\ref{table:ablation-study} details the results of our ablation study separately for each domain.

\begin{table*}[!tb]
	\centering
	\scalebox{0.9}{
		\begin{tabular}{lrrrrrrrrrrr|r}
			\hline
			\rowcolor{LightGray}
			                        & ECB           & EME           & Epp           & GNO           & JRC           & KDE           & \makebox[\widthof{ECB}]{News} & PHP           & TED           & Ubu           & Wiki          & \textbf{all}  \\ \hline
			\textit{~~test-0.4}                                                                                                                                                                                                                     \\ \hline
			\mlevt{3}               & 53.9          & \textbf{55.6} & \textbf{34.2} & 60.7          & \textbf{66.0} & \textbf{53.5} & \textbf{20.4}                 & 33.0          & \textbf{28.6} & 47.5          & \textbf{32.8} & \textbf{46.5} \\ \hline
			~~\texttt{-sel    }     & \textbf{54.5} & \textbf{55.6} & 32.7          & \textbf{61.2} & 65.9          & 52.2          & 19.5                          & \textbf{33.3} & 27.7          & \textbf{48.1} & 31.3          & 46.2          \\ \hdashline
			~~\texttt{-delx   }     & 52.2          & 53.4    ,     & 31.8          & 58.7          & 64.0          & 52.0          & 19.6                          & 31.2          & 27.5          & 46.8          & 31.5          & 44.8          \\ \hdashline
			~~\texttt{-rd-del }     & 49.7          & 47.6          & 22.2          & 48.2          & 56.7          & 38.8          & 13.2                          & 29.5          & 16.9          & 32.2          & 21.4          & 38.6          \\ \hdashline
			~~\texttt{-mask   }     & 53.4          & 54.7          & 33.7          & 58.9          & 65.3          & 52.4          & 20.3                          & 33.1          & 27.5          & 47.1          & 32.2          & 46.0          \\ \hdashline
			~~\texttt{-dum-plh}     & 50.8          & 43.3          & 32.7          & 45.6          & 61.2          & 42.4          & 21.0                          & 30.9          & 24.3          & 35.6          & 28.2          & 41.0          \\ \hdashline
			~~\texttt{-indep-align} & 51.5          & 52.4          & 29.7          & 53.8          & 60.7          & 47.2          & 16.9                          & 30.4          & 21.8          & 39.0          & 28.4          & 42.6          \\ \hline      \hline
			\textit{~~test-0.6}                                                                                                                                                                                                                     \\
			\hline
			\mlevt{3}               & \textbf{64.2} & \textbf{68.0} & 49.4          & \textbf{73.0} & \textbf{76.4} & \textbf{60.1} & 21.2                          & \textbf{39.6} & 52.2          & \textbf{60.1} & 61.6          & \textbf{60.1} \\ \hline
			~~\texttt{-sel    }     & 63.8          & 67.5          & \textbf{50.0} & 71.2          & 76.3          & 60.0          & \textbf{21.5}                 & 39.1          & \textbf{54.2} & 59.7          & \textbf{62.2} & 60.0          \\ \hdashline
			~~\texttt{-delx   }     & 62.1          & 66.7          & 47.7          & 70.6          & 75.0          & 58.8          & 20.1                          & 37.9          & 53.5          & 58.4          & 60.9          & 58.6          \\ \hdashline
			~~\texttt{-rd-del }     & 58.4          & 59.6          & 39.7          & 59.0          & 67.6          & 47.7          & 16.9                          & 35.0          & 39.8          & 44.1          & 47.4          & 51.9          \\ \hdashline
			~~\texttt{-mask   }     & 63.1          & 65.3          & 49.1          & 69.4          & 74.2          & 58.2          & 21.8                          & 38.6          & 50.9          & 58.7          & 59.7          & 59.0          \\ \hdashline
			~~\texttt{-dum-plh}     & 57.3          & 55.5          & 44.9          & 51.8          & 68.5          & 44.5          & 20.2                          & 35.7          & 42.6          & 38.7          & 50.7          & 50.9          \\ \hdashline
			~~\texttt{-indep-align} & 60.6          & 64.0          & 46.6          & 64.3          & 71.9          & 55.5          & 18.6                          & 35.6          & 44.6          & 51.6          & 56.0          & 56.4          \\ \hline
		\end{tabular}
	}
	\caption{\label{table:ablation-study} Ablation study. We report BLEU scores for various settings. \texttt{-sel}: no random selection noise ($\gamma$=0); \texttt{-delx}: no extra deletion loss;\texttt{-rd-del}: no random deletion ($\beta$=0); \texttt{-mask}: no random mask ($\delta$=0); \texttt{-dum-plh}: null probability to start with $\trg_{post \cdot del}$=$\trg_*$ ($\alpha=0$); \texttt{-indep-align}: the alignments are performed independently.
	}
\end{table*}

\paragraph{Illustration} A full inference run is in Table~\ref{table:illustration-php-04-571}, illustrating the benefits of considering multiple examples and realignment. Even though the realignment does not change here the final output, it reduces the number of atomic edits needed to generate it, making the inference more robust.

{
% \fontfamily{lmodern}\selectfont
\tiny
\begin{table*}[ht]
	\centering
	\scalebox{0.85}{
		\begin{tabular}{|rl|}
			\hline
			\textbf{src}: & The swf \_ setfont ( ) sets the current font to the value given by the fontid parameter .                                                                                                                                                                                                                  \\ \hline
			\textbf{tgt}: & swf \_ setfont ( ) remplace la police courante par la police repérée par l ' identifiant fontid .                                                                                                                                                                                                          \\ \hline\hline
			              & \textit{LevT}                                                                                                                                                                                                                                                                                              \\ \hline
			$\trg^{del}$: & sw$\bullet$	f	\_	\Ccancel{fon$\bullet$}	\Ccancel{t$\bullet$}	\Ccancel{size}	(	)	remplace	la	\Ccancel{taille}	de	\Ccancel{la}	police	par	la	\Ccancel{taille}	\Ccancel{size}	.                                                                                                                               \\ \hdashline
			$\trg^{plh}$: & sw$\bullet$	f	\_\textcolor{StrongGreen}{+2}	(	)	remplace	la\textcolor{StrongGreen}{+1}	de	police\textcolor{StrongGreen}{+1}	par	la\textcolor{StrongGreen}{+5}	.                                                                                                                                            \\ \hdashline
			$\trg^{tok}$: & sw$\bullet$	f	\_	\textcolor{StrongGreen}{set$\bullet$	police}	(	)	remplace	la	\textcolor{StrongGreen}{police}	de	police	\textcolor{StrongGreen}{courante}	par	la	\textcolor{StrongGreen}{valeur	de	paramètre	ti$\bullet$	d}	.                                                                                                       \\ \hdashline
			$\trg^{del}$: & sw$\bullet$	f	\_	set$\bullet$	police	(	)	remplace	la	police	de	police	courante	par	la	valeur	de	paramètre	ti$\bullet$	d	.                                                                                                                                                                                  \\ \hdashline
			$\trg^{plh}$: & \textcolor{StrongGreen}{+1}	sw$\bullet$	f	\_	set$\bullet$	police	(	)	remplace	la	police	de	police	courante\textcolor{StrongGreen}{+1}	par	la	valeur	de	paramètre\textcolor{StrongGreen}{+1}	ti$\bullet$	d	.                                                                                                \\ \hdashline
			$\trg^{tok}$: & \textcolor{StrongGreen}{Le}	sw$\bullet$	f	\_	set$\bullet$	police	(	)	remplace	la	police	de	police	courante	\textcolor{StrongGreen}{donnée}	par	la	valeur	de	paramètre	\textcolor{StrongGreen}{fon$\bullet$}	ti$\bullet$	d	.                                                                                \\ \hdashline
			\textbf{hyp}: & Le	\textcolor{red}{swf	\_}	setpolice	(	)	\textcolor{red}{remplace	la}	police	\textcolor{red}{de	police}	courante	donnée	\textcolor{red}{par	la}	valeur	de	paramètre	fontid	\textcolor{red}{.}                                                                                                                                                                                                   \\
			\hline
			\hline
			              & \mlevt{2}                                                                                                                                                                                                                                                                                           \\ \hline
			$\trg^{del}$:
			              & sw$\bullet$ f \_ \Ccancel{fon$\bullet$} \Ccancel{t$\bullet$} \Ccancel{size} ( ) remplace la \Ccancel{taille} \Ccancel{de} \Ccancel{la} police par \Ccancel{la} \Ccancel{taille} \Ccancel{size} .                                                                                                                                                                                                     \\
			              & sw$\bullet$ f \_ \Ccancel{defin$\bullet$} \Ccancel{e$\bullet$} font ( ) définit la police \Ccancel{fon$\bullet$} \Ccancel{t$\bullet$} \Ccancel{name} \Ccancel{et} \Ccancel{lui} \Ccancel{affecte} \Ccancel{l} \Ccancel{'} \Ccancel{identi$\bullet$} \Ccancel{fiant} fon$\bullet$ ti$\bullet$ d .                                                                                                                                   \\ \hdashline

			$\trg^{plh}$:
			              & sw$\bullet$ f \_\textcolor{StrongGreen}{+2} ( ) remplace la police\textcolor{StrongGreen}{+6} par\textcolor{StrongGreen}{+4} .                                                                                                                                                                                                                                                        \\
			              & sw$\bullet$ f \_\textcolor{StrongGreen}{+1} font ( ) définit la police\textcolor{StrongGreen}{+9} fon$\bullet$ ti$\bullet$ d .                                                                                                                                                                                                                               \\ \hdashline

			$\trg^{cmb}$:
			              & \Cu{sw$\bullet$} \Cu{f} \Cu{\_} \hphantom{set$\bullet$ font}  \Cu{(} \Cu{)} \Cu{remplace} \Cu{la} \Cu{police} \hphantom{actuelle à à la valeur donnée} \Cu{par} \hphantom{le paramètre fon$\bullet$ ti$\bullet$} \Ccancel{.}                                                                                                                                                                                                                                                  \\
			              & sw$\bullet$ f \_ \hphantom{set$\bullet$} \Cu{font} ( ) \makebox[\widthof{remplace}]{définit} la police \hphantom{actuelle à à la valeur donnée par le paramètre} \Cu{fon$\bullet$} \Cu{ti$\bullet$} \Cu{d} \Cu{.}                                                                                                                                                                                                                         \\ \hdashline

			$\trg^{tok}$:
			              & sw$\bullet$ f \_ set$\bullet$ font ( ) remplace la police \textcolor{StrongGreen}{actuelle à à la valeur donnée} par \textcolor{StrongGreen}{le paramètre} fon$\bullet$ ti$\bullet$ d .                                                                                                                                                                      \\ \hdashline

			$\trg^{del}$:
			              & sw$\bullet$ f \_ set$\bullet$ font ( ) remplace la police actuelle \Ccancel{à} \Ccancel{à} la valeur donnée par le paramètre fon$\bullet$ ti$\bullet$ d .                                                                                                                                                                      \\ \hdashline

			$\trg^{plh}$:
			              & sw$\bullet$ f \_ set$\bullet$ font ( ) remplace la police actuelle\textcolor{StrongGreen}{+1} la valeur donnée par le paramètre fon$\bullet$ ti$\bullet$ d .                                                                                                                                                                        \\ \hdashline
			$\trg^{tok}$:
			              & sw$\bullet$ f \_ set$\bullet$ font ( ) remplace la police actuelle \textcolor{StrongGreen}{à} la valeur donnée par le paramètre fon$\bullet$ ti$\bullet$ d .                                                                                                                                                                        \\ \hdashline
			\textbf{hyp}: & \textcolor{red}{swf \_} set\textcolor{red}{font ( ) remplace la police} actuelle à la valeur donnée \textcolor{red}{par} le paramètre \textcolor{red}{fontid} \textcolor{red}{.} \\
			\hline
			\hline
			              & \mlevt{3}                                                                                                                                                                                                                                                                                           \\ \hline
			$\trg^{del}$: & sw$\bullet$	f	\_	\Ccancel{fon$\bullet$}	\Ccancel{t$\bullet$}	\Ccancel{size}	(	)	remp$\bullet$ place	la	\Ccancel{taille}	\Ccancel{de}	\Ccancel{la}	police	par	\Ccancel{la}	\Ccancel{taille}	\Ccancel{size}	.                                                                                                \\
			              & sw$\bullet$	f	\_	\Ccancel{defin$\bullet$}	\Ccancel{e$\bullet$}	font	(	)	définit	la	police	\Ccancel{fon$\bullet$}	\Ccancel{t$\bullet$}	\Ccancel{name}	\Ccancel{et}	\Ccancel{lui}	\Ccancel{affecte}	\Ccancel{l}	\Ccancel{'}	\Ccancel{identi$\bullet$}	\Ccancel{fiant}	\Ccancel{fon$\bullet$}	ti$\bullet$	d	. \\
			              & sw$\bullet$	\Ccancel{f$\bullet$}	\Ccancel{text}	-	set$\bullet$	font	(	)	remplace	la	police	courante	par	\Ccancel{font}	.                                                                                                                                                                                   \\	\hdashline
			$\trg^{plh}$: & sw$\bullet$	f	\_\textcolor{StrongGreen}{+2}	(	)	remplace	la	police\textcolor{StrongGreen}{+4}	par\textcolor{StrongGreen}{+5}	.                                                                                                                                                                             \\
			              & sw$\bullet$	f	\_\textcolor{StrongGreen}{+1}	font	(	)	définit	la	police\textcolor{StrongGreen}{+7}	ti$\bullet$	d	.                                                                                                                                                                                          \\
			              & sw$\bullet$\textcolor{StrongGreen}{+2}	set$\bullet$	font	(	)	remplace	la	police	courante\textcolor{StrongGreen}{+4}	par\textcolor{StrongGreen}{+5}	.                                                                                                                                                       \\
			\hdashline
			$\trg^{cmb}$: & \Cu{sw$\bullet$}	\Cu{f}	\Cu{\_}	\hphantom{set$\bullet$ font}		\Cu{(}	\Cu{)}	\Cu{remplace}	\Cu{la}	\Cu{police}	\hphantom{courante}	\hphantom{au valeur donnée}			\Cu{par}					\hphantom{par ti$\bullet$	d	.} \hphantom{fon}	\Cu{.}                                                                          \\
			              & sw$\bullet$	f	\_	\hphantom{set$\bullet$}	\Cu{font}	(	)	\makebox[\widthof{remplace}]{définit}	la	police	\hphantom{courante par par} \hphantom{au valeur donnée}							\hphantom{fon} \Cu{ti$\bullet$}	\Cu{d}	\Cu{.}                                                                                         \\
			              & sw$\bullet$	\hphantom{f \_}		\Cu{set$\bullet$}	font	(	)	remplace	la	police	\Cu{courante}	\hphantom{par} \hphantom{au valeur donnée}				\Ccancel{par} \hphantom{fon} \hphantom{ti$\bullet$	d	. .}						\Cu{.}                                                                                               \\
			\hdashline
			$\trg^{tok}$: & sw$\bullet$	f	\_	set$\bullet$	font	(	)	remplace	la	police	courante	\textcolor{StrongGreen}{au	valeur	donnée}	par	\textcolor{StrongGreen}{le	fon$\bullet$}	ti$\bullet$	d	.	.	.                                                                                                                              \\ \hdashline
			$\trg^{del}$: & sw$\bullet$	f	\_	set$\bullet$	font	(	)	remplace	la	police	courante	\Ccancel{au}	valeur	\Ccancel{donnée}	par	le	fon$\bullet$	ti$\bullet$	d	.	\Ccancel{.}	\Ccancel{.}                                                                                                                                        \\ \hdashline
			$\trg^{plh}$: & \textcolor{StrongGreen}{+1}	sw$\bullet$	f	\_	set$\bullet$	font	(	)	remplace	la	police	courante\textcolor{StrongGreen}{+2}	valeur\textcolor{StrongGreen}{+1}	par	le\textcolor{StrongGreen}{+1}	fon$\bullet$	ti$\bullet$	d	.                                                                                 \\ \hdashline
			$\trg^{tok}$: & Le	sw$\bullet$	f	\_	set$\bullet$	font	(	)	remplace	la	police	courante	\textcolor{StrongGreen}{à	la}	valeur	\textcolor{StrongGreen}{donnée}	par	le	\textcolor{StrongGreen}{paramètre}	fon$\bullet$	ti$\bullet$	d	.                                                                                          \\ \hdashline
			\textbf{hyp}: & Le	\textcolor{red}{swf	\_	setfont	(	)	remplace	la	police	courante}	à	la	valeur	donnée		\textcolor{red}{par}	le	paramètre	fon\textcolor{red}{tid}	\textcolor{red}{.}                                                                                                                                                                                                              \\ \hline\hline 
			              & \mlevt{3} + \textit{realign}                                                                                                                                                                                                                                                                        \\ \hline
			$\trg^{del}$: & sw$\bullet$	f	\_	\Ccancel{fon$\bullet$}	\Ccancel{t$\bullet$}	\Ccancel{size}	(	)	remp$\bullet$ place	la	\Ccancel{taille}	\Ccancel{de}	\Ccancel{la}	police	par	\Ccancel{la}	\Ccancel{taille}	\Ccancel{size}	.                                                                                                \\
			              & sw$\bullet$	f	\_	\Ccancel{defin$\bullet$}	\Ccancel{e$\bullet$}	font	(	)	définit	la	police	\Ccancel{fon$\bullet$}	\Ccancel{t$\bullet$}	\Ccancel{name}	\Ccancel{et}	\Ccancel{lui}	\Ccancel{affecte}	\Ccancel{l}	\Ccancel{'}	\Ccancel{identi$\bullet$}	\Ccancel{fiant}	\Ccancel{fon$\bullet$}	ti$\bullet$	d	. \\
			              & sw$\bullet$	\Ccancel{f$\bullet$}	\Ccancel{text}	-	set$\bullet$	font	(	)	remplace	la	police	courante	par	\Ccancel{font}	.                                                                                                                                                                                   \\	\hdashline
			$\trg^{plh}$: & sw$\bullet$	f	\_\textcolor{StrongGreen}{+2}	(	)	remplace	la	police\textcolor{StrongGreen}{+4}	par\textcolor{StrongGreen}{+5}	.                                                                                                                                                                             \\
			              & sw$\bullet$	f	\_\textcolor{StrongGreen}{+1}	font	(	)	définit	la	police\textcolor{StrongGreen}{+8}	ti$\bullet$	d	.                                                                                                                                                                                          \\
			              & sw$\bullet$\textcolor{StrongGreen}{+2}	set$\bullet$	font	(	)	remplace	la	police	courante\textcolor{StrongGreen}{+3}	par\textcolor{StrongGreen}{+5}	.                                                                                                                                                       \\
			\hdashline
			$\trg^{cmb}$: & \Cu{sw$\bullet$}	\Cu{f}	\Cu{\_}	\hphantom{set$\bullet$ font} \Cu{(}	\Cu{)}	\Cu{remplace}	\Cu{la}	\Cu{police}	\hphantom{courante}	\hphantom{au valeur donnée} \Cu{par} \hphantom{le paramètre fon$\bullet$ ti$\bullet$	d} \Cu{.}                                                                            \\
			              & sw$\bullet$	f \_ \hphantom{set$\bullet$} \Cu{font} ( ) \makebox[\widthof{remplace}]{définit} la police \hphantom{courante au valeur donnée par le paramètre fon$\bullet$} \Cu{ti$\bullet$} \Cu{d} .                                                                                                        \\
			              & sw$\bullet$	\hphantom{f \_} \Cu{set$\bullet$}	font (	) remplace	la	police	\Cu{courante} \hphantom{au valeur donnée} par \hphantom{le paramètre fon$\bullet$ ti$\bullet$	d} .                                                                                                                               \\
			\hdashline
			$\trg^{tok}$: & sw$\bullet$	f \_ set$\bullet$ font ( ) remplace la	police	courante \textcolor{StrongGreen}{au	valeur	donnée}	par	\textcolor{StrongGreen}{le paramètre fon$\bullet$}	ti$\bullet$ d .                                                                                                                        \\ \hdashline
			$\trg^{del}$: & sw$\bullet$	f	\_	set$\bullet$	font	(	)	remplace	la	police	courante	\Ccancel{au}	valeur donnée	par	le paramètre	fon$\bullet$	ti$\bullet$	d	.                                                                                                                                                                \\ \hdashline
			$\trg^{plh}$: & \textcolor{StrongGreen}{+1}	sw$\bullet$	f	\_	set$\bullet$	font	(	)	remplace	la	police	courante\textcolor{StrongGreen}{+2}	valeur	donnée par	le	paramètre fon$\bullet$	ti$\bullet$	d	.                                                                                                                      \\ \hdashline
			$\trg^{tok}$: & Le	sw$\bullet$	f	\_	set$\bullet$	font	(	)	remplace	la	police	courante	\textcolor{StrongGreen}{à	la}	valeur	donnée	par	le	paramètre	fon$\bullet$	ti$\bullet$	d	.                                                                                                                                            \\ \hdashline
			\textbf{hyp}: & Le	\textcolor{red}{swf	\_	setfont	(	)	remplace	la	police	courante}	à	la	valeur	donnée		\textcolor{red}{par}	le	paramètre	fon\textcolor{red}{tid}	\textcolor{red}{.}                                                                                                                                                                                                             \\ \hline
		\end{tabular}
	}
	\caption{\label{table:illustration-php-04-571} Examples of full inference of several models on a test sample from \textit{test-0.4-PHP} (sample n°571). Copied parts are in red.}
\end{table*}
}
% \hphantom

\end{document}